\documentclass[pdflatex,sn-mathphys-num,iicol]{sn-jnl} 
\usepackage{graphicx}%
\usepackage{multirow}%
\usepackage{amsmath,amssymb,amsfonts}%
\usepackage{amsthm}%
\usepackage{mathrsfs}%
\usepackage[title]{appendix}%
\usepackage{xcolor}%
\usepackage[most]{tcolorbox}
\tcbuselibrary{listings}

\newtcblisting{bashbox}[1]{
  colback=black!5,
  colframe=black!70,
  listing only,
  sharp corners,
  title=#1,
  listing options={
    language=bash,
    basicstyle=\ttfamily\small,
    breaklines=true,
    columns=fullflexible
  }
}

\newtcblisting{pythonbox}[1]{
  colback=black!5,
  colframe=black!70,  
  listing only,
  sharp corners,
  title=#1,
  fonttitle=\bfseries\small,
  listing options={
    language=Python,
    basicstyle=\ttfamily\footnotesize,
    breaklines=true,
    columns=fullflexible,
    commentstyle=\color{green!50!black},
    keywordstyle=\color{blue!80!black},
    stringstyle=\color{orange!80!black},
    showstringspaces=false
  }
}

\usepackage{textcomp}%
\usepackage{manyfoot}%
\usepackage{booktabs}%
\usepackage{algorithm}%
\usepackage{algorithmicx}%
\usepackage{algpseudocode}%
\usepackage{listings}%
\usepackage[title]{appendix}
\usepackage[normalem]{ulem} 
\usepackage{pifont}
\newcommand{\cmark}{\ding{51}}%
\newcommand{\xmark}{\ding{55}}%
\usepackage{fontawesome}
\usepackage{arydshln}
\usepackage{caption}
\theoremstyle{thmstyleone}%

\theoremstyle{thmstyletwo}%

\theoremstyle{thmstylethree}%

\begin{document}

\title[Article Title]{Bharat Scene Text: A Novel Comprehensive Dataset and Benchmark for Indian Language Scene Text Understanding}

\author{\fnm{Anik} \sur{De}}
\author{\fnm{Abhirama Subramanyam} \sur{Penamakuri}}
\author{\fnm{Rajeev} \sur{Yadav}}
\author{\fnm{Aditya} \sur{Rathore}}
\author{\fnm{Harshiv} \sur{Shah}}
\author{\fnm{Devesh} \sur{Sharma}}
\author{\fnm{Sagar} \sur{Agarwal}}
\author{\fnm{Pravin} \sur{Kumar}}
\author*{\fnm{Anand} \sur{Mishra}} \email{mishra@iitj.ac.in}

\affil[1]{\orgname{Indian Institute of Technology Jodhpur}, \orgaddress{\city{Jodhpur}, \postcode{342030}, \state{Rajasthan}, \country{India}} \\\url{https://vl2g.github.io/projects/IndicPhotoOCR/}}

\abstract{Reading scene text, that is, text appearing in images, has numerous application areas, including assistive technology, search, and e-commerce. Although scene text recognition in English has advanced significantly and is often considered nearly a solved problem, Indian language scene text recognition remains an open challenge. This is due to script diversity, non-standard fonts, and varying writing styles, and, more importantly, the lack of high-quality datasets and open-source models.

To address these gaps, we introduce the Bharat Scene Text Dataset (BSTD) -- a large-scale and comprehensive benchmark for studying Indian Language Scene Text Recognition. It comprises more than 100K words that span 11 Indian languages and English, sourced from over 6,582 scene images captured across various linguistic regions of India. The dataset is meticulously annotated and supports multiple scene text tasks, including: (i) Scene Text Detection, (ii) Script Identification, (iii) Cropped Word Recognition, and (iv) End-to-End Scene Text Recognition.
We evaluated state-of-the-art models originally developed for English by adapting (fine-tuning) them for Indian languages. Our results highlight the challenges and opportunities in Indian language scene text recognition. We believe that our dataset represents a significant step toward advancing research in this domain. All our models and data are open source.}


\maketitle
\begin{figure}[!t]
    \centering
    \includegraphics[width=1\linewidth]{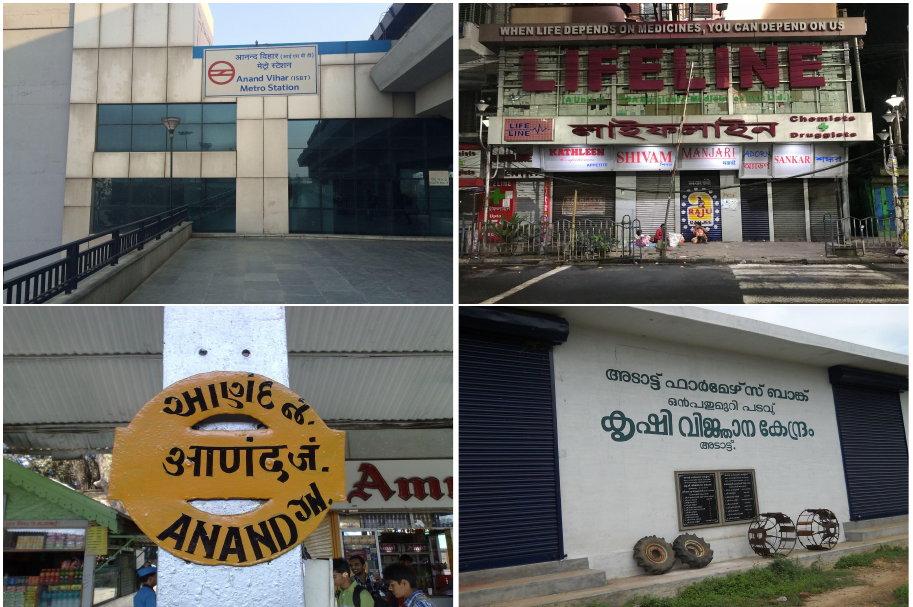}
    \caption{\textbf{Scene Text Understanding: Case of India.} This figure showcases typical street scenes from northern, eastern, western, and southern parts of India, highlighting the country's vast linguistic diversity. With 11 languages including English commonly featured on signboards, scene text understanding in India presents unique challenges. Scene Text from five languages, namely Hindi, English, Bengali Gujarati and Tamil are shown here. While significant progress has been made in English Scene Text Recognition, \emph{open-source comprehensive effort} for Indian language scene text understanding, including large public datasets and open-source models, are still limited. Our work seeks to address this gap toward advancing the field.}
    \label{fig:goal}
\end{figure}
\section{Introduction}\label{sec1}

Scene Text Recognition, especially for English, has advanced significantly in the last decade, thanks to a series of efforts on dataset creation~\cite{ICDAR2003, SVT, mishra2012scene, MSRA-TD500, ICDAR2013, ICDAR2015, CUTE80,Total-Text, SVT-P, MLT-19} 
and model development~\cite{cnn_detector_2014, CRNN2016,jaderberg2016reading,  east,  textspotter2018, craft, ASTER2019, ABCNet2020, dbnet, visionLAN2021, ABINet2021, parseq, ABINet++2023, textbpn++, clip4str}.
In contrast, scene text recognition for Indian languages has been extremely underexplored and possibly more challenging (refer to Fig.~\ref{fig:goal}). Note that Indian languages are used by more than 1.4 billion people, which is approximately 18\% of the global population. Despite its usage and popularity, we observe that scene text recognition in Indian languages lacks a comprehensive benchmark and open-source models, except for some early efforts~\cite{recognition-ISI-umapada2010, IIIT_ILST, detection-ISI-Umapada2017,kakwani-etal-2020-indicnlpsuite, IIIT-TL-STR, IIIT-indicSTR12, VijayanCDK25} that laid important groundwork, but did not aim for a comprehensive coverage of tasks or languages. We aim to fill this gap through our work.

To accelerate advancements in Indian Language Scene Text Recognition, we present a comprehensive benchmark called the \emph{\underline{B}harat \underline{S}cene \underline{T}ext \underline{D}ataset} (BSTD)\footnote{India, traditionally known as \emph{Bharat}, reflects the nation's rich cultural and linguistic diversity, which inspired the naming of our dataset.}. The BSTD comprises 6.5K scene images that feature more than 100K words in 11 Indian languages, along with English. These images that are sourced from Wikimedia\footnote{\url{https://www.wikimedia.org/}}, span various signboards and accidental background text from different Indian cities and towns, ensuring that the dataset captures the linguistic diversity of India. For each visible text instance in an image, we manually annotate a polygonal bounding region and provide its corresponding transcription. The dataset is publicly available for download\footnote{\url{https://github.com/Bhashini-IITJ/BharatSceneTextDataset}}.

The BSTD can be used for the following scene text tasks: (i) scene text detection, (ii) script identification, (iii) cropped word recognition, and (iv) end-to-end scene text recognition. In this work, we benchmark the BSTD using our new pipeline, \raisebox{-0.19\height}{\includegraphics[height=1.2em]{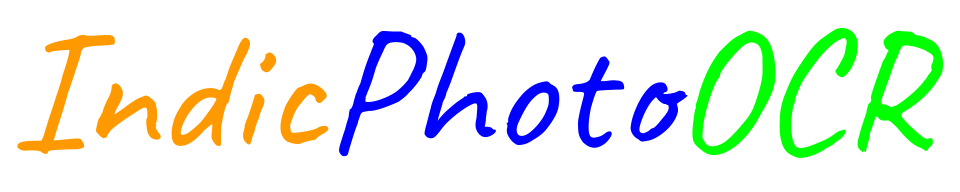}}, which integrates state-of-the-art architectures in its modules. We firmly believe that these benchmarks will help accelerate advancements in Indian language scene text recognition. 

\begin{table*}[!t]
\centering
\caption{\textbf{Comparison of the Bharat Scene Text Dataset (BSTD) with existing scene text datasets}. BSTD supports all three core tasks - text detection, script identification, and recognition (and, thereby end-to-end evaluation) across 11 Indian languages along with English. It is the \textbf{only publicly available dataset for Indian languages that enables comprehensive end-to-end scene text understanding}. \textit{$^*$: MLT-17 and MLT-19 are not focused on Indian languages, they only include Hindi and Bengali. $^{\$}$: IIIT-IndicSTR12 and $^{\#}$: IIIT-STR~\cite{GunnaSJ22} are primarily annotated for cropped word recognition;} consequently, neither can be utilized to evaluate end-to-end scene text recognition as ours.}
\label{tab:dataComp}
\resizebox{1\textwidth}{!}
{
\begin{tabular}{lccccccc}
\toprule
\multirow{2}{*}{Dataset} & \multirow{2}{*}{\# Images} & \multirow{2}{*}{\# Words} & \multirow{2}{*}{\# Languages} & \multicolumn{3}{c}{Tasks Supported} \\
\cmidrule(lr){5-7}
& & & & Detection & Script Identification & Recognition \\
\midrule
\midrule
ICDAR 2003~\cite{ICDAR2003}       & 509     & 2,268    & 1  & \cmark & \xmark & \cmark \\
SVT~\cite{SVT}                    & 350     & 725     & 1  & \cmark & \xmark & \cmark \\
IIIT-5K~\cite{mishra2012scene}& -     & 5,000  & 1 & \xmark & \xmark & \cmark \\
ICDAR 2013~\cite{ICDAR2013}       & 462     & 5,003    & 1  & \cmark & \xmark & \cmark \\
ICDAR 2015~\cite{ICDAR2015}       & 1,500    & 6,545    & 1  & \cmark & \xmark & \cmark \\
CUTE80~\cite{CUTE80}              & 80      & 288     & 1  & \cmark & \xmark & \cmark \\
Total-Text~\cite{Total-Text}      & 1,555    & 11,459   & 1  & \cmark & \xmark & \cmark \\
IIIT-TL-STR~\cite{IIIT-TL-STR}    & 440     & 3,035    & 2  & \xmark & \xmark & \cmark \\
IIIT-ILST~\cite{IIIT_ILST}        & -       & 3,168    & 3  & \xmark & \xmark & \cmark \\
MLT-17$^*$~\cite{MLT-17}              & 18,000  & 96,000  & 9  & \cmark & \cmark & \cmark \\
MLT-19$^*$~\cite{MLT-19}              & 20,000  & 191,000 & 10 & \cmark & \cmark & \cmark \\
IIIT-indicSTR12$^{\$}$~\cite{IIIT-indicSTR12} & -   & 27,000  & 12 & \xmark & \xmark & \cmark \\
IIIT-STR$^{\#}$~\cite{GunnaSJ22} &  -  & 12,057  & 6 & \xmark & \xmark & \cmark \\
\textbf{BSTD (This Work)}                       & 6,582    & 1,26,292  & 12 & \cmark & \cmark & \cmark \\
\bottomrule
\end{tabular}
}
\end{table*}

\begin{figure}[!t]
    \centering
    \includegraphics[height=8cm]{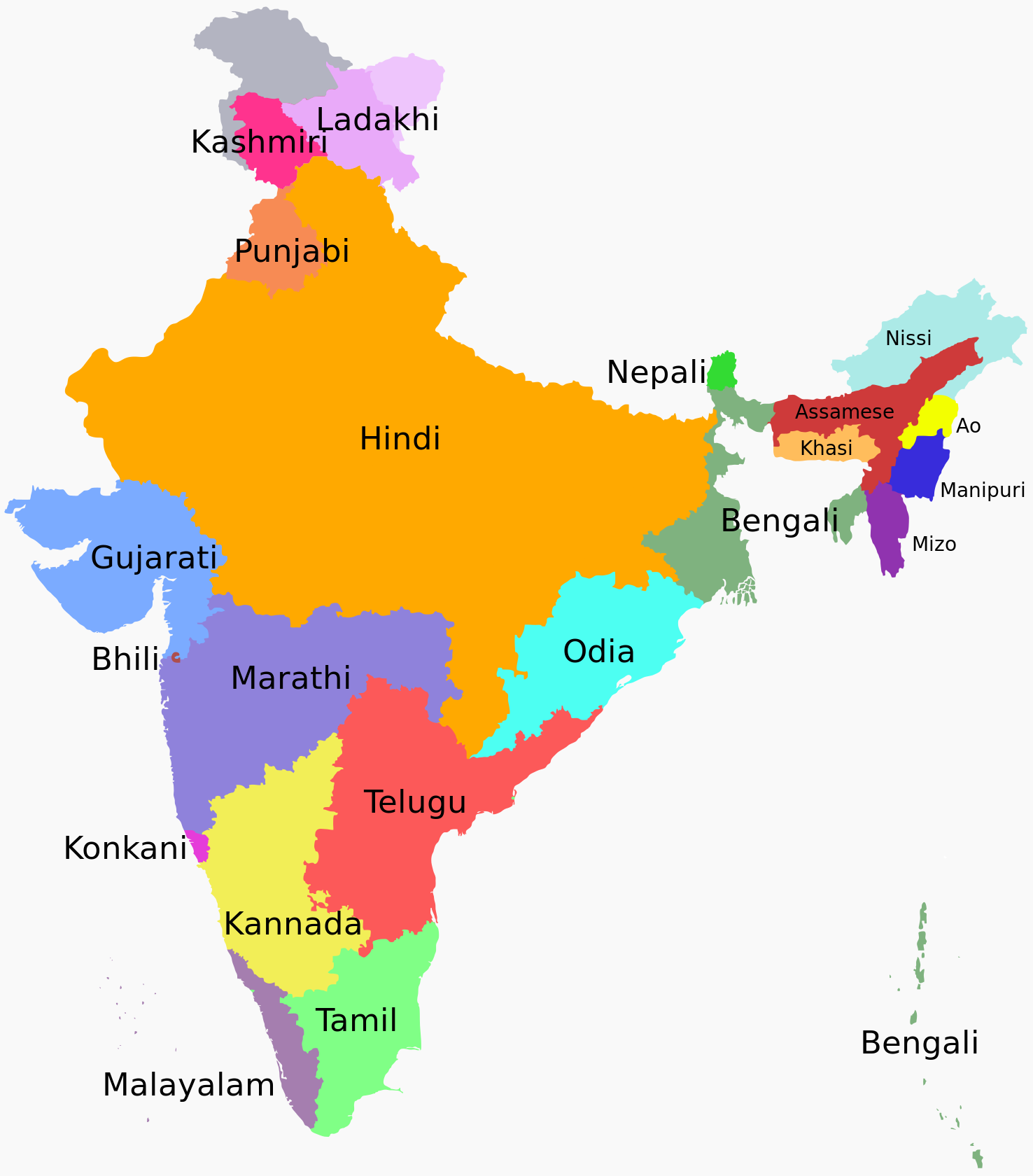}
    \caption{Language map of India (Source: \href{https://commons.wikimedia.org/wiki/File:Language_region_maps_of_India.svg}{Wikimedia}). The BSTD covers scene text images from 11 prominent languages namely Assamese, Bengali, Gujarati, Hindi, Kannada, Malayalam, Marathi, Odia, Punjabi, Tamil and, Telugu. Additionally, it contains English, as English is often included as one of the languages as signboard in India.}
    \label{fig:indian_language_map}
\end{figure}

We list our contributions here:
\begin{enumerate}
    \item We introduce the Bharat Scene Text Dataset (BSTD) -- a large-scale benchmark for Indian language scene text recognition, consisting of 6,582 scene images featuring 1,26,292 words in 11 Indian languages and English. Each image is manually annotated with polygon-level bounding boxes and corresponding transcription and script, ensuring high-quality data for research and applications. (Section~\ref{sec:bstd})

    \item We present a strong baseline and release corresponding open-source toolkit, namely \raisebox{-0.19\height}{\includegraphics[height=1.2em]{images/IndicPhotoOCR_LOGO.png}}, designed to detect, identify, and recognize text in English and 11 Indian languages. The released toolkit is easy to install and replicate, making it accessible to researchers and developers. (Section~\ref{sec:ipo} and Appendix-A)

    \item We evaluated BSTD using state-of-the-art techniques, including off-the-shelf methods and fine-tuned models on synthetic and/or BSTD train data (as applicable). Detailed benchmark results including failure case analysis are provided for all key tasks, i.e., text detection, script identification, cropped word recognition, and end-to-end scene text recognition. These evaluations establish a strong baseline for future research and advancements in multilingual scene text recognition, particularly for Indian languages. (Section~\ref{sec:eval})
\end{enumerate}

\begin{figure*}[!t]
\centering
  \includegraphics[width=\textwidth]{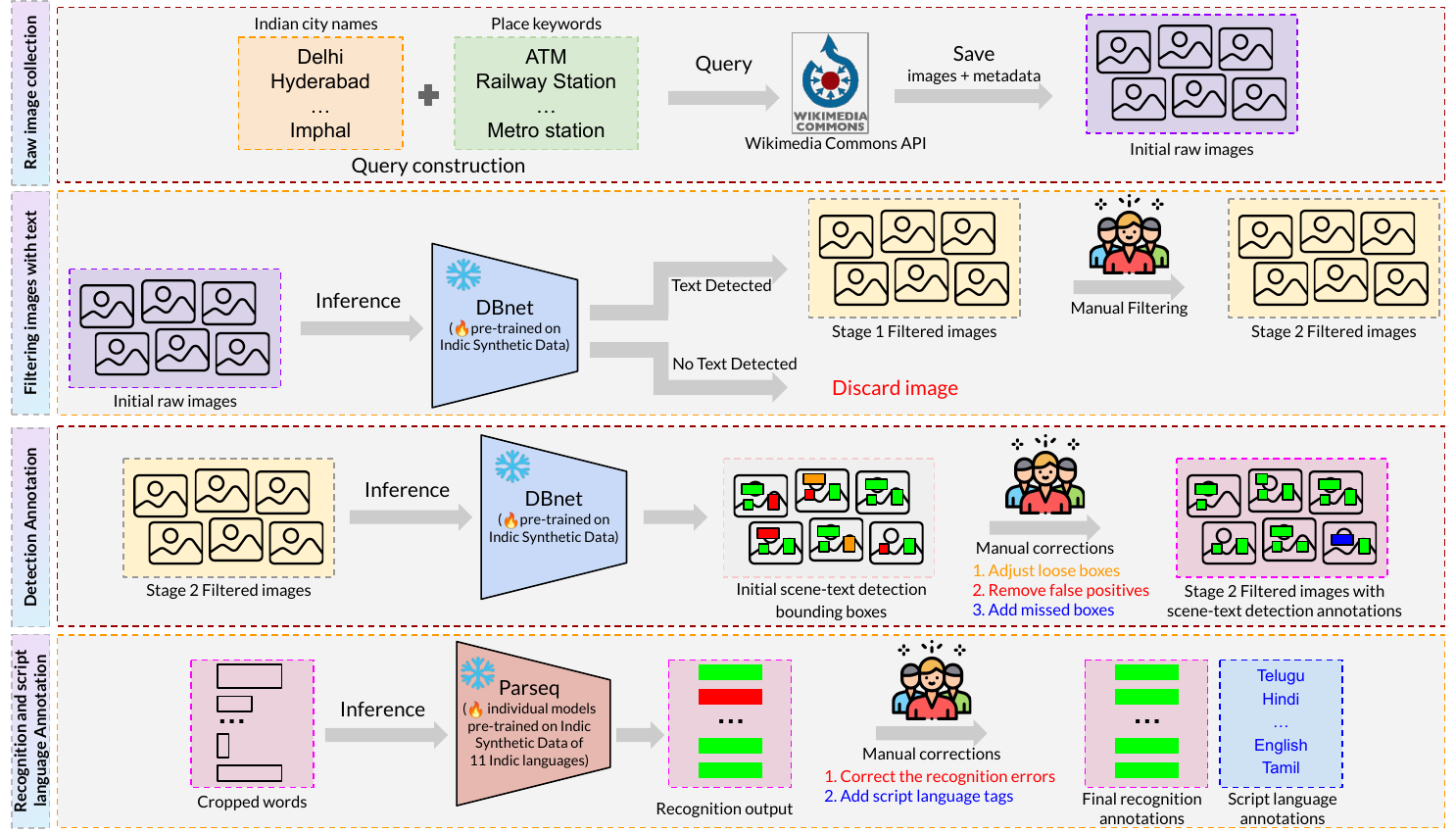}
  \caption{Pipeline for constructing a multilingual scene-text dataset from Wikimedia Commons images. The process begins with query construction using combinations of Indian city names and place-related keywords to retrieve relevant images via the Wikimedia Commons API. A two-stage filtering process using DBnet~\cite{dbnet} (pre-trained on Indic Synthetic Data) removes non-textual images and manually filters relevant ones. Scene-text detection annotations are generated using DBnet and refined through manual corrections. Finally, cropped word images are recognized using the PARSeq~\cite{parseq} model (trained on 11 Indic languages using synthetic data), with recognition errors corrected and script language tags added to produce the final annotated dataset.}
  \label{fig:bstd_curation_pipeline}
\end{figure*}

\section{Bharat Scene Text Dataset}
\label{sec:bstd}
India, often referred to as Bharat (the Indian name for the country), represents a rich and diverse cultural and linguistic landscape, as illustrated in Fig.~\ref{fig:indian_language_map}. In the context of scene text understanding, the vast multilingual environment poses unique challenges, as the scripts, writing styles, and text formats vary significantly between different regions. In addition, the identification of scripts becomes extremely important. Although there has been some effort in collecting and benchmarking Indian language scene text understanding, such as~\cite{IIIT_ILST,IIIT-TL-STR, IIIT-indicSTR12, GunnaSJ22}, it falls short compared to efforts for scene text understanding in English~\cite{ICDAR2003, SVT, mishra2012scene, MSRA-TD500, ICDAR2013, ICDAR2015, CUTE80,Total-Text, SVT-P, MLT-19}. To address this gap, we present the Bharat Scene Text Dataset, a novel and comprehensive dataset specifically curated to represent the linguistic diversity of India. By using the term ``Bharat", we not only acknowledge the nation's deep-rooted cultural identity but also highlight our focus on supporting scene text understanding across 11 major Indian languages. This dataset contains 6,582 scene images and 1,26,292 word images in total. This dataset and associated annotations can be downloaded from our project website\footnote{\url{https://vl2g.github.io/projects/IndicPhotoOCR/}}. Please refer to Table~\ref{tab:dataComp} for a detailed comparison of the Bharat Scene Text Dataset compared to other related scene text datasets. 

In the next section, we give details of dataset curation and annotation, supported tasks, train-test splits followed by dataset analysis.

\subsection{Dataset Curation and Annotation}
We illustrate the pipeline for constructing the Bharat Scene Text Dataset from Wikimedia Commons in Fig.~\ref{fig:bstd_curation_pipeline}.
The dataset is curated and annotated in the following three stages:

\subsubsection{Stage-1: Image collection} We compile a comprehensive list of cities and towns across various regions of India where 11 selected languages are spoken predominantly. Furthermore, we curate a catalog of public places, such as `bus stations', `banks', `schools', `railway stations', `hospitals', `ATMs', `metro stations', `street signs', `airport', `shop entrance', `billboards', 'press conference', etc. By combining the city and place names as search queries, e.g. `Delhi Railway Station', `Kolkata Bookshop', we retrieved images along with their metadata from the Wikimedia Commons API\footnote{\url{https://commons.wikimedia.org/w/api.php}}. It should be noted that not all retrieved images contain scene text, necessitating a subsequent filtering process. At this stage, we obtain $\approx$ 50K images. 

\begin{figure*}[!t]
    \centering
    \includegraphics[width=15cm]{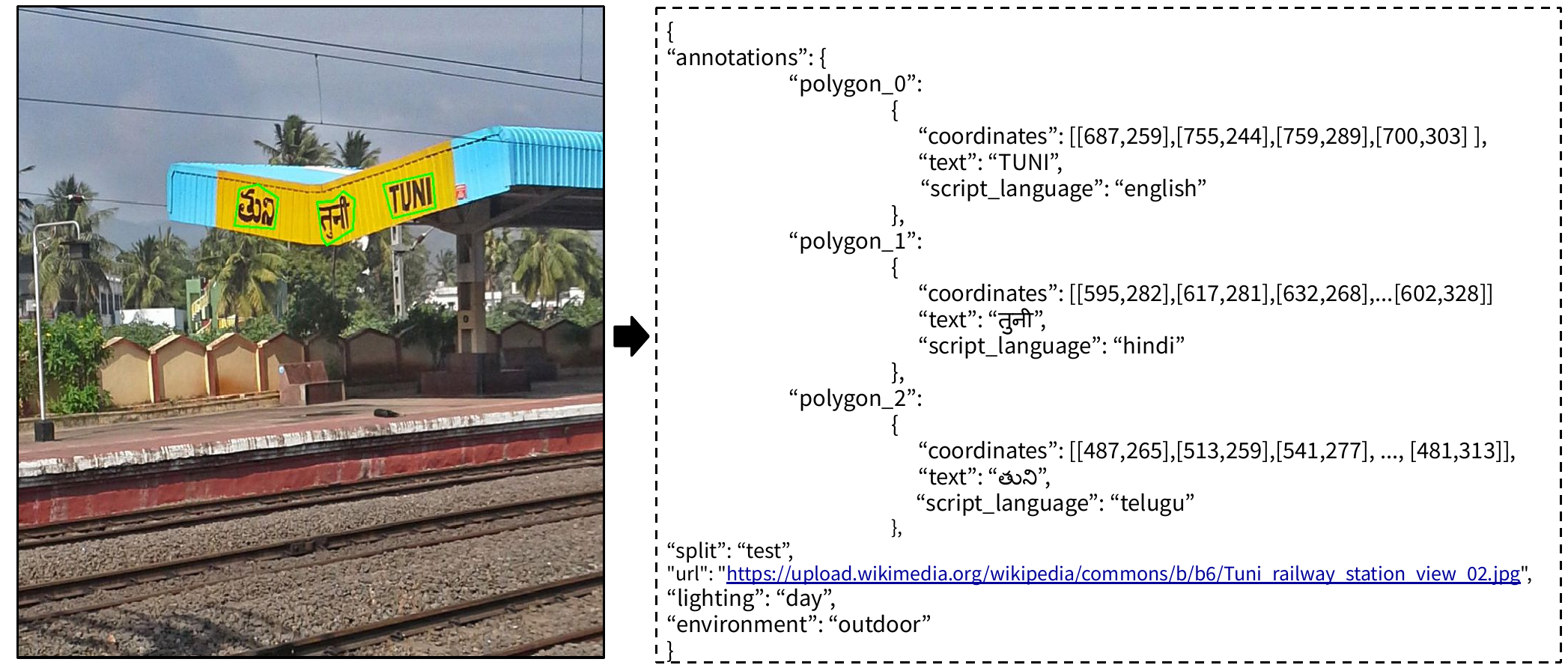}
    \caption{An example image from BSTD along with the corresponding JSON annotation. }
    \label{fig:bstd_annotation}  
\end{figure*}

\subsubsection{Stage-2: Image filtering}
\label{subsec:image_filter}
We filter the images harvested in the previous step using the pre-trained scene text detection method~\cite{dbnet}. To this end, we run this detector and obtain text even with low confidence. We retain images containing at least one word,  while others were discarded. Nevertheless, 53 images in the final dataset were found to contain no annotatable scene text. We deliberately retained this small fraction of non-text images to preserve the natural variability of real-world data collection pipelines, where some captured scenes may not ultimately contain visible text. The final data set comprises 6,582 images that contain scene text in 11 Indian languages and English.

\subsubsection{Stage-3: Semi-automatic annotation}
\label{subsec:detect_annot}
We annotate the final filtered images for scene text detection in two steps to reduce human annotation efforts.  We obtain bounding box annotation using a fine-tuned DBnet model on synthetic Indian language data\footnote{Refer to Section-\ref{sec:sdg} for details of synthetic data generation}. Since these automatically obtained bounding boxes are not always perfect, they were refined by our hired annotators using an internal tool to edit, draw, or delete polygons around the scene text words. 

Once the bounding boxes were obtained, they were assigned to the annotators and tasked to label each box with the script it contains. These annotators were well-versed in script identification and were also informed about the geographical origin of each image. With this context, they were able to accurately label each bounding box with theier corresponding script. We then use PARSeq~\cite{parseq}, fine-tuned on synthetic data for Indian languages, to read the text within each bounding box. The bounding boxes, along with the recognized text, are then provided to language expert annotators. These annotators review the text and correct any recognition errors or retain the text as is if it is accurate. In addition, for every image, we have included annotation on the lighting (captured during the day or night) and environment (captured outdoor or indoor) conditions.

Fig.~\ref{fig:bstd_annotation} shows the fine grained nature of the annotations done in this work.

\subsection{Annotator Details}
We employed two distinct groups of annotators for this work. Initially, three human annotators for image filtering (Section~\ref{subsec:image_filter}) and bounding box annotation (Section~\ref{subsec:detect_annot}). For the next two steps, text correction and script assignment (refer~\ref{subsec:detect_annot}), we hired eleven language-specific experts, one for each target language. The annotators were proficient in reading, understanding, and writing their native language, as well as English. They were shown cropped words and asked to correct the pseudo-recognition annotations and assign the script. All final annotations underwent an additional quality check to eliminate potential errors and noise. The annotators were compensated at rates aligned with standard wages for a tier-2 city in India. The entire process--pseudo-annotation, manual correction, and quality verification took 10 months.

\subsection{Supported Tasks and Evaluation Protocols}
The Bharat Scene Text Dataset (BSTD) enables us to study the following four tasks:
\begin{enumerate}
    \item \textbf{Scene Text Detection}: where given an image, the task is to detect all the text instances in the image along with their bounding boxes. We provide polygon-level tight bounding boxes for each word image in the dataset. The popular metrics such as Precision, Recall and F1-score using TedEval evaluation~\cite{tedeval} can be used to assess the performance of the scene text detectors.

    \item \textbf{Script Identification}: This task involves identifying the language of the script present in a given cropped word image. The performance of the model is evaluated using accuracy. Additionally, we include a confusion matrix to analyze where the model's predictions fail.

    \item  \textbf{Cropped Word Recognition}: where the task is to recognize and extract the text given a cropped word image provided the script/language of text is known. For cropped word recognition, the word recognition rate (WRR) is used as a performance metric. 

    \item \textbf{End-to-End Scene Text Recognition}: This task is the closest to the natural setting. Here, given a scene image, the task is to recognize all instances of text. Solving end-to-end Scene Text Recognition for Indian languages, often requires a pipeline of scene text detection, script identification followed by cropped word recognition. We use mean word recognition rate (WRR) and character recognition rate (CRR) over all the test images in the dataset. The WRR and CRR for an image are defined as follows:
    
    \begin{equation}
    \text{WRR} = \left( 1 - \frac{S_{w} + D_{w} + I_{w}}{N_{w}^{\text{total}}} \right) \times 100
    \end{equation}
    
    \begin{equation}
    \text{CRR} = \left( 1 - \frac{S_{c} + D_{c} + I_{c}}{N_{c}^{\text{total}}} \right) \times 100
    \end{equation}

    where $S$, $D$, and $I$ with subscript $w$ and $c$ denote the number of substitutions, deletions, and insertions, respectively, at the word or character level. Further, $N_{w}^{\text{total}}$ and $N_{c}^{\text{total}}$ denote the total number of words and characters present in the ground-truth transcription in the image, respectively. It should be noted that due to false alarms (i.e., excessive insertions of words or characters in the recognized text) or in some rare cases of any missing annotation, the computed WRR or CRR values can become negative for few images. To keep the interpretation simple, we truncate such values to 0. Furthermore, standard WRR and CRR computations assume a fixed reading order; however, since reading order in scene text images can be subjective, we relax this constraint in our evaluation.  
    
    Further, to make interpretation of results more clean, we also use standard precision, recall, and F1-score to evaluate end-to-end performance. Note that precision is defined as the number of correctly recognized words over the total number of recognized words, while recall is defined as the number of correctly recognized words over the total number of words in the ground truth. 
\end{enumerate}

\begin{table}[!t]
\centering
\small
\caption{Summary of bounding box annotation in the BSTD, detailing the total, training, and test set distributions} 
\label{tab:detectionDataSplit} 
\begin{tabular*}{\columnwidth}{@{\extracolsep{\fill}}lccc}
\toprule
Set & Images & Bounding Box Annotations \\ 
\midrule
\midrule
Total & 6,582 & 1,26,292 \\
Train & 5,263 & 94,128 \\ 
Test & 1,319 & 32,164 \\ 
\bottomrule
\end{tabular*}
\end{table}

We randomly divided the 6,582 images into an 80-20\% train-test split. As a result, the training set consists of 5,263 scene images and 77,726 cropped words, each with corresponding language and text annotations. The test set contains 1,319 scene images and 28,752 cropped words with their respective language and text annotations. The image-wise and cropped world-level train/test split is summarized in Table~\ref{tab:detectionDataSplit} and Table~\ref{tab:recognitionDataSplit}, respectively. For the script identification task, we randomly selected an equal number of word images from each of the 12 languages. The split comprises 1,800 training images and 478 testing images per language.

\begin{figure}[!t]
    \centering
    \includegraphics[width=\columnwidth]{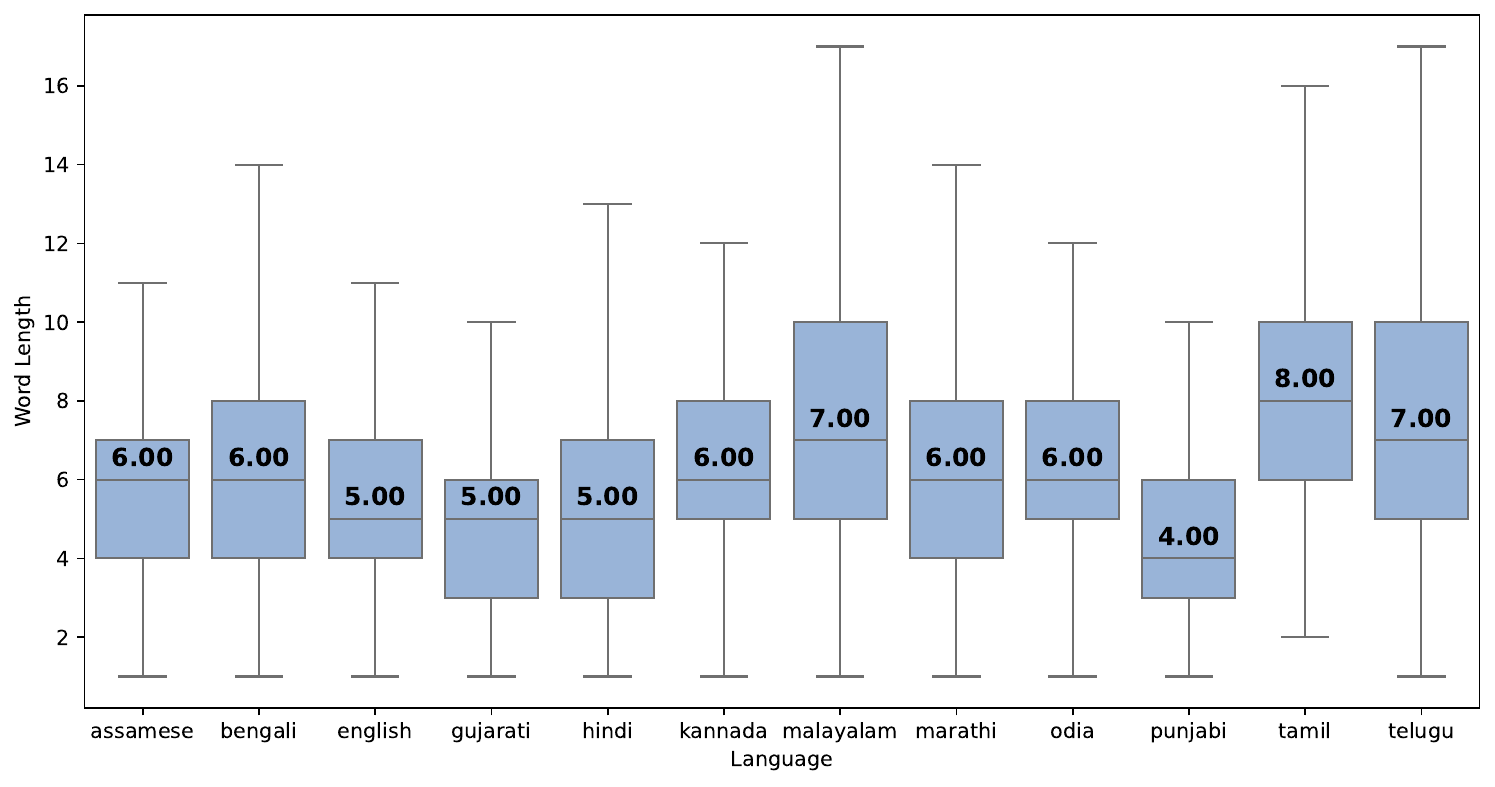}  
    \caption{Box plot highlighting the word length range across all the 12 languages.}
    \label{fig:boxplot_wordlength}
\end{figure}

\begin{table}[!t]
\centering
\small
\caption{Cropped word image data split for the BSTD, showing the number of annotated instances, train images, and test images across multiple languages. Here, Others correspond to those words which are not from the listed languages here (e.g., Meitei and Urdu). We plan to include their annotation in subsequent versions of our dataset. For scene-level language distribution statistics, please refer to Table~\ref{tab:image_level_language_distribution}.}
\label{tab:recognitionDataSplit}
\begin{tabular*}{\columnwidth}{@{\extracolsep{\fill}}lccc}
\toprule
Language &  \#Word images & Train & Test \\
\midrule
\midrule
Assamese & 4,132 & 2,627 & 1,505 \\ 
Bengali & 6,304 & 4,936 & 1,368 \\
English & 41,696 & 29,123 & 12,573 \\
Gujarati & 2,899 & 1,884 & 1,015 \\
Hindi & 19,773 & 14,927 & 4,846 \\
Kannada & 2,928 & 2,208 & 720 \\
Malayalam & 2,940 & 2,393 & 547 \\
Marathi & 5,113 & 3,917 & 1,196 \\
Odia & 4,192 & 3,148 & 1,044 \\
Punjabi & 11,199 & 8,319 & 2,880 \\
Tamil & 2,542 & 2,029 & 513 \\
Telugu & 2,760 & 2,215 & 545 \\
Others & 19,814 & - & - \\
\midrule
Total & 1,26,292 & 77,726 & 28,752 \\
\bottomrule
\end{tabular*}
\end{table}

\begin{figure}[!t]
    \centering
    \includegraphics[width=\columnwidth]{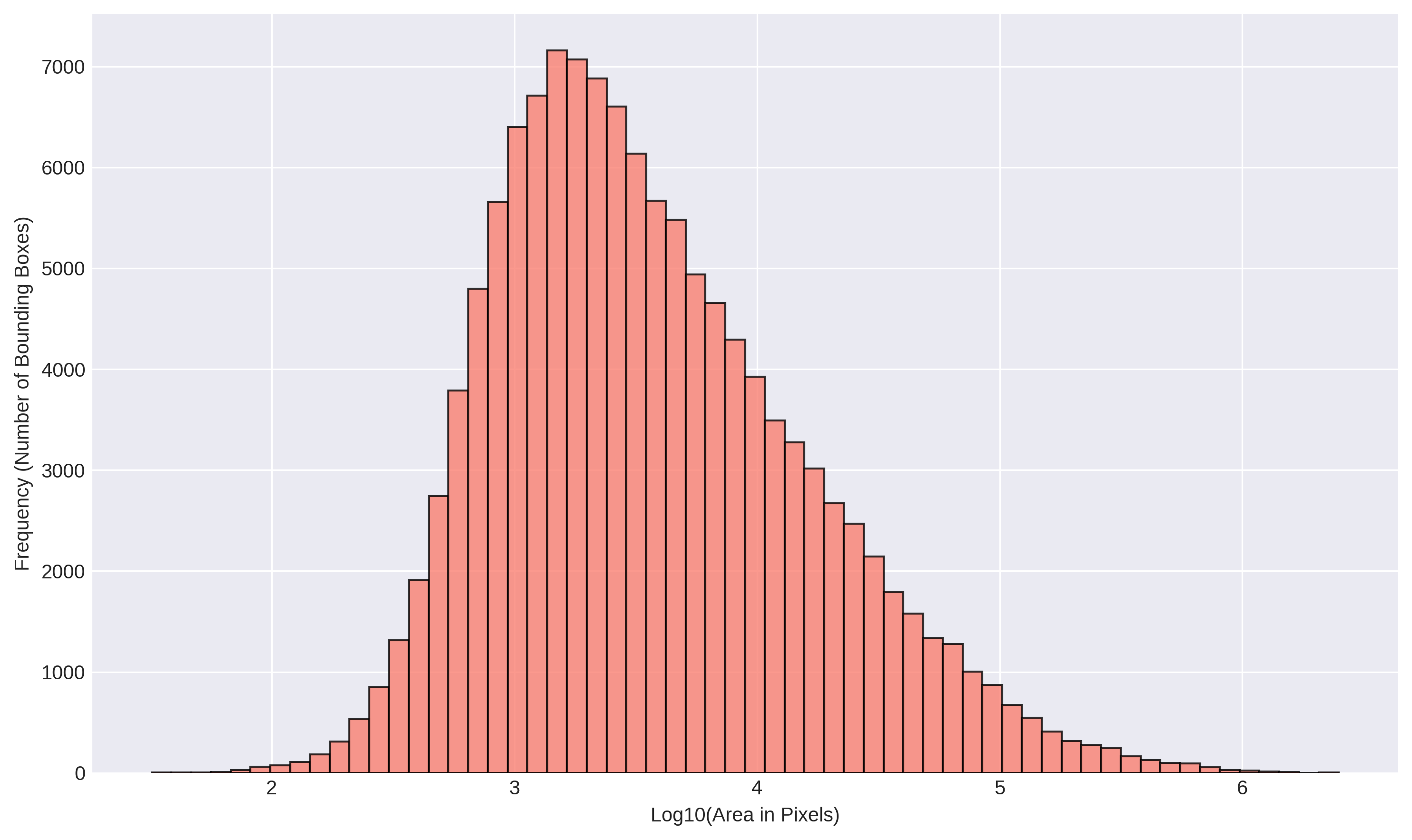}
    \caption{Distribution of word-level bounding box areas in the BSTD, plotted on a base-10 logarithmic scale.}
    \label{fig:detectionbox_area_distribution}  
\end{figure}

\begin{table}[!t]
\centering
\small
\caption{Scene-level language composition of the BSTD. The distribution highlights the multilingual nature of the dataset, with a significant majority of images containing English co-occurring with one or more Indian languages.}
\label{tab:image_level_language_distribution}
\begin{tabular*}{\columnwidth}{@{\extracolsep{\fill}}lr} 
\toprule
Language Composition & \#Images \\
\midrule
English only & 53 \\
Indian languages only & 1,958 \\
English + one or more Indian languages & 4,518 \\
No annotatable scene text & 53 \\
\midrule
Total & 6,582 \\
\bottomrule
\end{tabular*}
\end{table}

\subsection{Dataset Analysis}
\label{sec:dataset_analysis}
In this section, we present a comprehensive analysis of the Bharat Scene Text Dataset (BSTD) to evaluate its quality and the challenges it poses. Toward this end, we first report the language-wise word count in Table~\ref{tab:recognitionDataSplit}. We observe that it has a good coverage of all the major Indian languages with minimum 2,542 word images for Tamil. Further, we performed an image-level analysis and observed that, out of the 6,582 total scene images in the dataset, only 53 contain exclusively English text, 1,958 contain only non-English text (i.e., Indian languages), and 4,518 contain English co-occurring with one or more Indian languages. This makes the dataset linguistically rich (refer to Table~\ref{tab:image_level_language_distribution}). We further analyze the average word length for each language in our dataset in Fig.~\ref{fig:boxplot_wordlength}. The plot reveals that Malayalam, Tamil, and Telugu contain longer words on average, whereas English and Hindi have shorter word lengths. This observation, though potentially intuitive, holds practical implications for the development of specialized language models in the future.

Moving further, we studied the distribution of the bounding box area in Fig.~\ref{fig:detectionbox_area_distribution}. It indicates that bounding box sizes range from very small text region, e.g., minimum area = 32 pixels to very-large text region, e.g., maximum area $\approx$ $2.49 \times 10^6$ pixels. The distribution closely approximates a log-normal curve that peaks near the median area of 2,889 pixels. These findings highlight that the BSTD contains a mix of small and large text regions within images, making it challenging for detection. Furthermore, the high proportion of small text regions indicates the prevalence of \emph{incidental text}, as opposed to only focused text, thereby making the dataset more challenging and representative of real-world scenarios.


An analysis of the lighting and environmental conditions of the scene images revealed that approximately 89\% of the images were captured during daylight, whereas the remaining 11\% were taken at night. Further, the environment annotations indicate that 91\% of the images correspond to outdoor scenes, while only 9\% were captured indoors.


In short, the analysis of the Bharat Scene Text Dataset suggests that it has diverse linguistic and geographical coverage, with a clear focus on text from public, commercial and transport-related signage, and has a mix of focused and incidental text, making it a valuable resource for developing robust real-world text recognition models for Indian languages. Further, while widely spoken languages such as Hindi, English, Bengali, and Marathi are well represented in the dataset, low-resource languages including Assamese, Gujarati, Kannada, Malayalam, Odia, Punjabi, Tamil, and Telugu also have a substantial presence, with at least 2,542 word images for Tamil (see Table~\ref{tab:recognitionDataSplit}). This shows that the dataset is not only strong in high-resource languages, but also sufficiently representative of low-resource languages, making it suitable for comprehensive multilingual text recognition research.

\section[IndicPhotoOCR: A Competitive Proposed Baseline]%
{\raisebox{-0.19\height}{\includegraphics[height=1.5em]{images/IndicPhotoOCR_LOGO.png}}: Proposed Baseline}
\label{sec:ipo}
Scene text recognition in a multilingual environment typically involves three key modules: (i) Scene Text Detection, (ii) Script Identification, and (iii) Cropped Word Recognition. In the absence of a dedicated architecture for Indian language scene text recognition in the literature, we develop a strong baseline by adapting or fine-tuning (as needed) state-of-the-art architectures across these modules. We describe each of these modules below and collectively refer to this pipeline as \raisebox{-0.19\height}{\includegraphics[height=1.5em]{images/IndicPhotoOCR_LOGO.png}}.   

\subsection{Scene Text Detection}
Our systematic analysis of modern scene text detectors revealed that these models are largely language-agnostic. Hence, we directly incorporate them into our pipeline without language-specific modifications. In particular, we adopt TextBPN++~\cite{textbpn++} for scene text detection, as it demonstrated superior performance compared to other detectors in our evaluations. Detailed results for this stage are presented in the experimental section~\ref{sec:detection}.

\subsection{Script Identification}
For script classification, we used the \texttt{google/vit-base-patch16-224-in21k} Vision Transformer (ViT) model, which consists of 12 transformer layers and uses $16 \times 16$ patch embeddings. Pre-trained on the large-scale ImageNet-21k dataset, this model provides rich visual representations that are well-suited for transfer learning for the task of script identification.

To adapt the model for our task, we appended a custom classification head that projects the learned visual features into output logits, corresponding to either 12 classes (for full multilingual classification) or 3 classes (a practical setting with Hindi, English, and one regional language at a time).

We fine-tuned the model, including both the transformer backbone and the classification head, using the AdamW optimizer with a learning rate of $2 \times 10^{-4}$ and a batch size of 16 using BSTD-train set.  This ViT-based approach consistently outperformed other configurations, demonstrating strong generalization in both 3-way and 12-way script classification settings.

\begin{table}[t!]
\centering
\caption{Statistics for Synthetic Dataset generated using~\cite{synthtext} for 11 Indian languages.}
\label{tab:synth_dataset_stats}
\setlength{\tabcolsep}{3.5pt} 
\footnotesize 
\begin{tabular}{l r r r r}
    \toprule
    \textbf{Language} & \textbf{Train Size} & \textbf{Test Size} & \textbf{Vocab Size} & \textbf{\#Fonts} \\
    \midrule
    Assamese  & 8,092,205 & 899,134 & 636,312   & 44  \\
    Bengali   & 8,475,805 & 941,757 & 4,447,622 & 26  \\
    Gujarati  & 6,431,896 & 714,656 & 4,255,185 & 57  \\
    Hindi     & 8,357,906 & 811,976 & 4,195,486 & 101 \\
    Kannada   & 4,002,547 & 444,728 & 8,288,718 & 22  \\
    Malayalam & 7,199,420 & 400,000 & 4,857,185 & 32  \\
    Marathi   & 3,168,386 & 352,043 & 3,739,734 & 101 \\
    Odia      & 4,041,852 & 449,095 & 1,144,724 & 35  \\
    Punjabi   & 5,479,762 & 608,863 & 2,083,909 & 40  \\
    Tamil     & 5,652,718 & 628,080 & 7,842,880 & 70  \\
    Telugu    & 3,485,570 & 387,286 & 5,705,371 & 42  \\
    \bottomrule
\end{tabular}
\end{table}

\subsection{Cropped Word Recognition}
Once text instances are detected and their scripts identified, the next step is text recognition. In contrast to detection, this task depends strongly on language-specific modeling and demands large-scale training data. \emph{Existing off-the-shelf English scene text recognition models are not directly applicable, as they are not designed for Indian scripts.} To address this, we fine-tune a successful scene text recognition architecture, PARSeq~\cite{parseq}, for the 11 Indian languages included in our study. A key challenge in this setting is the scarcity of real-world scene text data for low-resource languages, in contrast to English, where recognizers are trained on millions of samples. To address this, we generated synthetic word images to augment the training data and bridge this gap as described in the following section.

\subsubsection{Synthetic Data Generation}
\label{sec:sdg}
To train models for each of the 11 languages, we required a large-scale dataset of scene text images. However, such datasets for Indian languages were not readily available. This lack of data led us to adopt an existing method for generating synthetic scene text images. We used SynthText~\cite{synthtext} to generate large-scale data, with statistics summarized in Table~\ref{tab:synth_dataset_stats}. SynthText~\cite{synthtext} provided 8000 background images, which we leveraged as the background image for our synthetic data. 

To generate scene text in different Indian languages, we needed appropriate vocabulary and fonts for each language. We sourced vocabulary from the AI4Bharat-IndicNLP Dataset\footnote{\label{ai4bharat}\url{https://github.com/AI4Bharat/indicnlp_corpus}} introduced in the IndicNLPSuite~\cite{kakwani-etal-2020-indicnlpsuite}. SynthText also accounts for local 3D geometry, enabling varied orientation of text placement within the scene. Although the generated images lacked complete realism, the scale of data generation offered significant value for our training needs.

This synthetically generated data is then used to train the PARSeq model for each language. PARSeq requires accurate character sets, which vary by language (e.g. Assamese: 91, Bengali: 91, Gujarati: 89; Hindi: 125, Kannada: 85, Mayalayam: 116, Marathi: 125, Odia: 90, Punjabi:75, Tamil: 64, Telugu: 98). 
During this stage, we used an input image size of $32 \times 128$, a maximum label length of 25, and random augmentation. Following synthetic training, the best model is then fine-tuned on real data (BSTD) using decaying learning rate applied layer-wise.

\section{Experimental Analysis}
\label{sec:eval}
In this section, we provide a detailed evaluation of the proposed baseline, namely \raisebox{-0.19\height}{\includegraphics[height=1.2em]{images/IndicPhotoOCR_LOGO.png}}, and its individual modules on the newly introduced Bharat Scene Text Dataset. We compare it with several open-source OCR systems, including Tesseract~\cite{tesseract}, PaddleOCR~\cite{cui2025paddleocr30technicalreport}, EasyOCR~\cite{jaidedai_easyocr_2020}, and SuryaOCR~\cite{paruchuri2025surya}. For end-to-end evaluation, we also compare it with commercial closed-source systems such as Google OCR~\cite{googlecloud_visionocr} and GPT-4~\cite{gpt_4v}. In addition, we include appropriate competitive approaches to evaluate individual modules. Our results reveal both the strengths and limitations of existing systems in handling diverse multilingual Indian scene text. They also demonstrate the effectiveness of \raisebox{-0.19\height}{\includegraphics[height=1.2em]{images/IndicPhotoOCR_LOGO.png}} in a challenging setting. Please note that all the quantitative results are reported on the Bharat Scene Text Dataset (test set).

\begin{table}[!t]
\centering
\caption{Scene Text Detection results on BSTD-Test Set using techniques prevalent in English scene text datasets. }
\label{tab:det_results}
\begin{tabular*}{\columnwidth}{@{\extracolsep{\fill}}lcccc}
\toprule
Method     & P & R & F1 \\ 
\midrule
\midrule
EAST~\cite{east}               & 0.55   & 0.10   & 0.17          \\ 
CRAFT~\cite{craft}             & 0.51   & 0.22   & 0.19          \\ 
DBNet~\cite{dbnet}             & 0.71   & 0.50   & 0.59          \\ 
Hi-SAM~\cite{Hi-SAM}           & \textbf{0.75}   & 0.19   & 0.46         \\ 
TextBPN++~\cite{textbpn++}     & \textbf{0.75}   & \textbf{0.78}   & \textbf{0.77}          \\ 
\bottomrule
\end{tabular*}
\end{table}

\subsection{Results on Scene-text detection}
\label{sec:detection}
Our proposed baseline, \raisebox{-0.19\height}{\includegraphics[height=1.2em]{images/IndicPhotoOCR_LOGO.png}}, employs TextBPN++~\cite{textbpn++} for scene text detection. We compare its performance with several widely used scene text detection methods, including EAST~\cite{east}, CRAFT~\cite{craft}, DBNet~\cite{dbnet}, and Hi-SAM~\cite{Hi-SAM}. The comparison results are summarized in Table~\ref{tab:det_results}. We observe that TextBPN++ significantly outperforms the other detectors, achieving higher F1-score and recall. Although Hi-SAM is also equally precise, it misses out detecting many Indian language words. Please note that we use TedEval~\cite{tedeval} evaluation in our experiments.
\begin{table}[t!]
\centering
\caption{Script identification accuracy (\%) in the 3-way classification setting: Hindi vs English vs one regional Indian language shown in the language column. Each row corresponds to a different regional Indian language, while Hindi and English remain fixed across all rows.}
\label{tab:script-id}
\small 
\begin{tabular*}{\columnwidth}{@{\extracolsep{\fill}}l c c c c }
\toprule
 & \multicolumn{4}{c}{Trained on BSTD} \\
 \cmidrule{2-5}
\textbf{Language} & AlexNet & CRNN & CLIP & Proposed \\ \midrule
 Assamese      & 53.1  & 72.0  & 75.6  & \textbf{92.2} \\
 Bengali       & 54.0  & 75.7  & 77.5  & \textbf{93.3} \\
 Kannada       & 54.7  & 73.8  & 82.7  & \textbf{94.5} \\
 Malayalam     & 54.3  & 71.6  & 84.1  & \textbf{90.8} \\
 Marathi       & 45.9  & 62.3  & 70.1  & \textbf{79.6} \\
 Odia          & 50.1  & 87.2  & 86.7  & \textbf{93.5} \\
 Tamil         & 51.3  & 75.3  & 79.6  & \textbf{93.6} \\
 Telugu        & 54.9  & 82.9  & 86.9  & \textbf{95.0} \\
 Gujarati      & 50.7  & 70.2  & 79.1  & \textbf{83.1} \\
 Punjabi       & 51.3  & 81.1  & 79.7  & \textbf{92.3} \\
\bottomrule
\end{tabular*}
\end{table}

\subsection{Results on Script Identification}
For script identification, we evaluate performance under two settings:
(i) \textbf{3-way classification}, where the model distinguishes among a regional language, Hindi and English. This setup reflects a common real-world scenario in India, where signboards often include all three languages.
(ii) \textbf{12-way classification}, where the model selects from the English and 11 Indian languages. 

We use the following baselines:
(i) AlexNet~\cite{alexnet}: finetuned on trainset of our script identification data, (ii) CRNN~\cite{CRNN2016}: a CNN-based scene text recognition model with the ability to encode sequence information, adapted for script identification, followed by state-of-the-art transformer architectures for image classification; (iii) CLIP~\cite{clip} and (iv) ViT~\cite{vit}, respectively. 

\begin{figure}[!t]
\centering
  \includegraphics[width=\columnwidth]{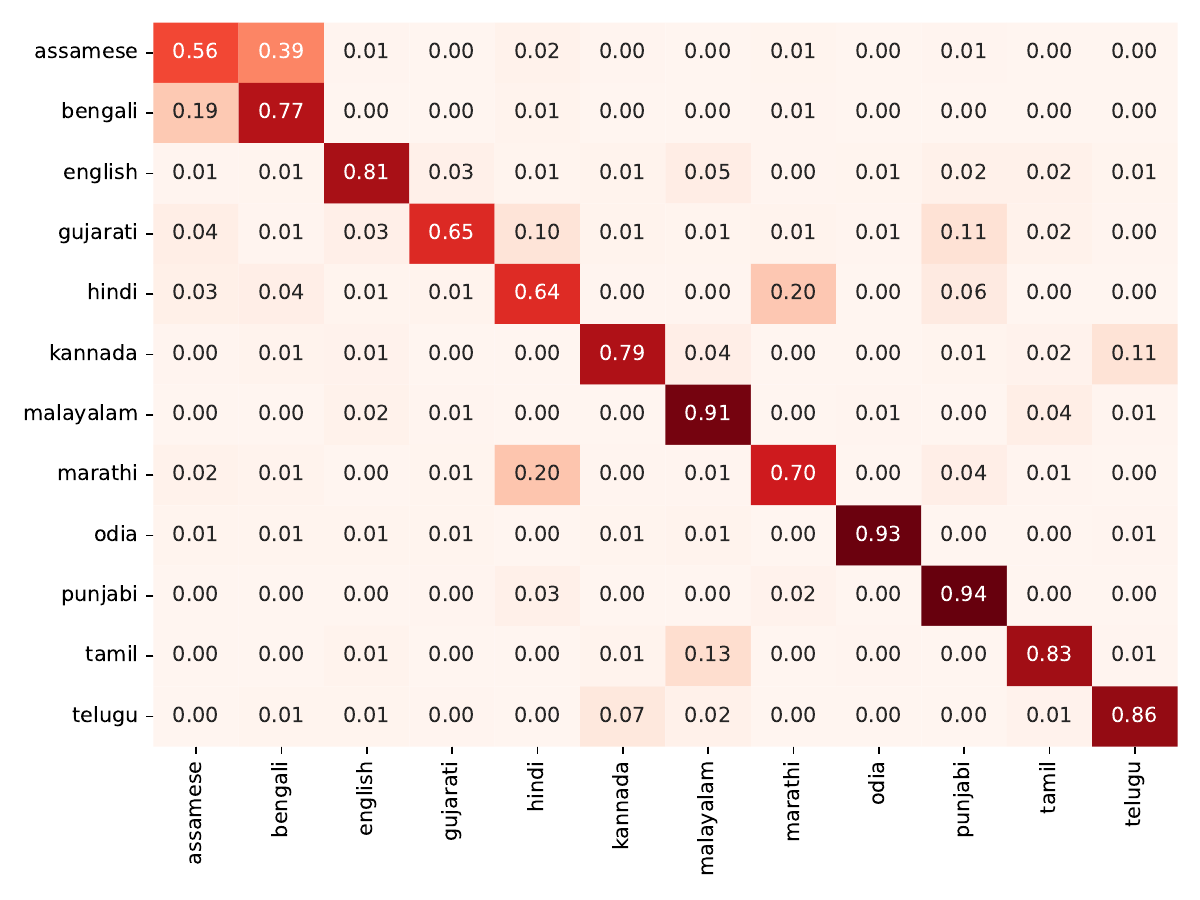}
    \caption{Normalized confusion matrix illustrating the performance of script identification in 12-way classification setting. \textbf{(Best viewed in color)}.}
  \label{fig:12-way_cf}
\end{figure}

For the 3-way classification, the results are reported in Table~\ref{tab:script-id}. We observe that VIT-based proposed script identification clearly outperforms other baselines. For 12-way script classification, the CLIP-based baseline achieves an accuracy of 67.7\%, while our proposed ViT-based approach achieves 80.5\%. The corresponding normalized confusion matrix is presented in Fig.~\ref{fig:12-way_cf}. While the overall results are satisfactory, the model frequently confuses Assamese with Bengali and Hindi with Marathi. This is primarily due to their sharing Unicode patterns. Also, since they share majority of the Unicodes, these two confusion does not affect significantly our end-to-end performance. We have also noticed that other languages frequently confuse with English, likely due to some data bias or weak context for small words (such as `\&', 'ON', etc.). Our end-to-end evaluation suggests that further improving script identification can significantly boost overall scene text recognition performance. 

\begin{table*}[!t]
\centering
\caption[Baseline results on cropped word recognition]{Baseline Results on Cropped Word Scene Text Recognition for 12 languages on BSTD. Please note that PARSeq is used in our proposed baseline -- \raisebox{-0.19\height}{\includegraphics[height=1.2em]{images/IndicPhotoOCR_LOGO.png}}.}
\label{tab:recog_results}
\resizebox{\textwidth}{!}{
\setlength{\tabcolsep}{4pt} 
\footnotesize 
\begin{tabular}{l c c c c c c c c c c c c c c} 
\toprule
Model & Train Data & Assamese & Bengali & English & Gujarati & Hindi & Kannada & Malayalam & Marathi & Odia & Punjabi & Tamil & Telugu & Average \\
\midrule
\multirow{2}{*}{PARSeq~\cite{parseq}} & Synth &0.47 &0.44 &0.92 &0.32 &0.43 &0.47 &0.52 &0.44 &0.42 &0.4 &0.37 &0.44 &0.47 \\
\cmidrule(l){2-15} 
& BSTD Train &0.79 &0.82 &0.92 &0.61 &0.71 &0.6 &0.58 &0.86 &0.72 &0.75 &0.8 &0.56 &0.73 \\
\midrule
CRNN~\cite{CRNN2016} & AI4Bharat &NA &NA &0.27 &NA &0.13 &NA &0.06 &NA &NA &0.27 &0.16 &0.05 &0.16 \\
\midrule
\multicolumn{15}{l}{\textbf{Off-the-shelf open source models}} \\ 
\midrule
Tesseract~\cite{tesseract} & - &0.15 &0.18 &0.33 &0.12 &0.2 &0.12 &0.03 &0.21 &0.11 &0.19 &0.13 &0.09 &0.15 \\
PaddleOCR~\cite{cui2025paddleocr30technicalreport} & - &NA &NA &0.61 &NA &0.31 &0.03 &NA &0.36 &NA &NA &0.28 &0.15 &0.29 \\
EasyOCR~\cite{jaidedai_easyocr_2020} & - &0.12 &0.24 &0.29 &NA &0.18 &0.06 &NA &0.25 &NA &NA &NA &0.11 &0.18 \\
\bottomrule
\end{tabular}}
\end{table*}

\begin{figure*}[!t]
\centering
  \includegraphics[width=16cm]{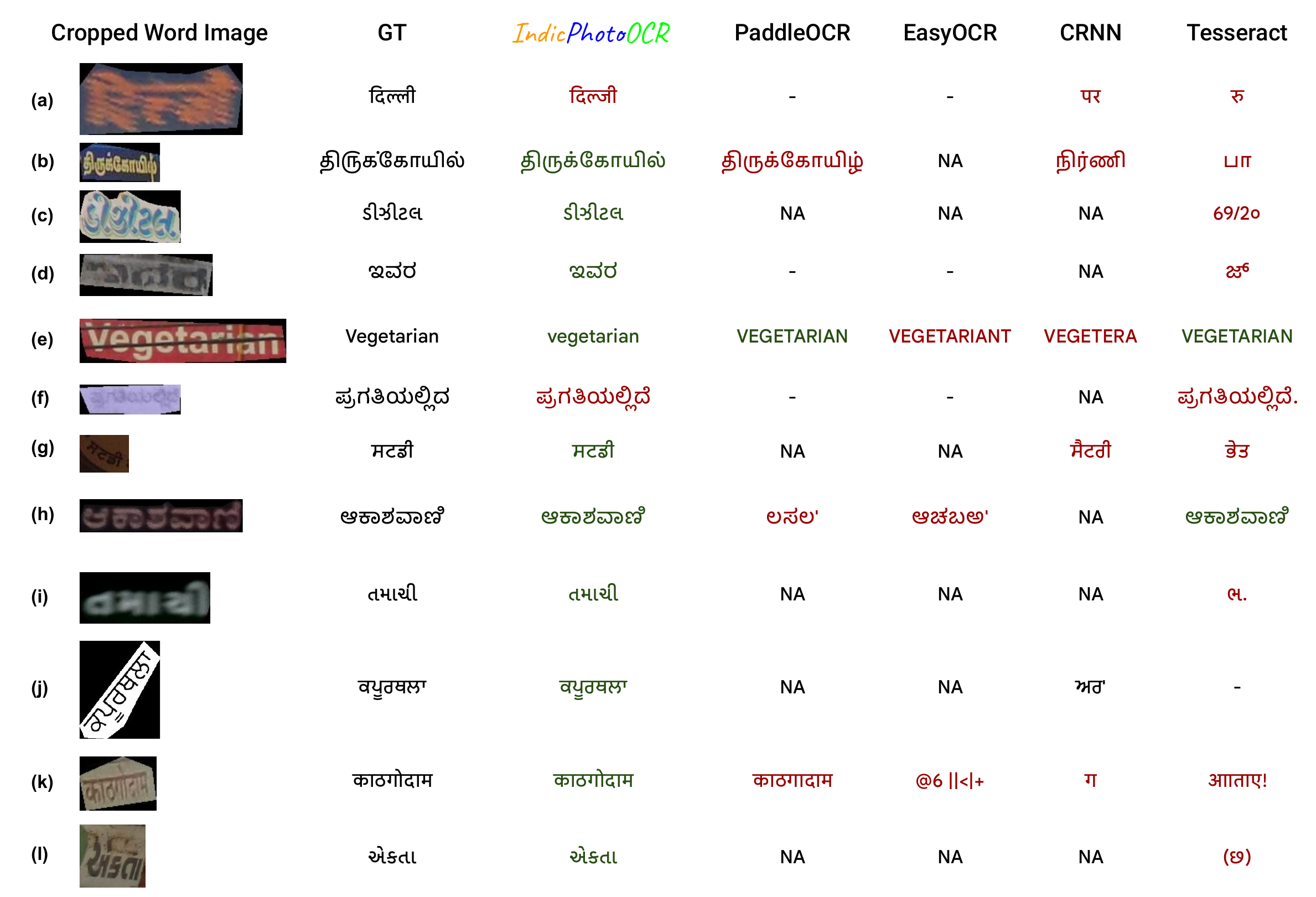}
  \caption[BSTD challenges with predictions]{\textbf{A selection of challenging cropped words from our dataset along with ground truth and recognized results by various methods including \raisebox{-0.19\height}{\includegraphics[height=1.2em]{images/IndicPhotoOCR_LOGO.png}}}. Key challenges include out-of-focus text (a, b, i), varied fonts (c), eroded text (d, k), handwritten text (f), occluded text (e, h, l), and skewed text (g, j). Cells marked with ``-" indicate an empty recognition, while ``NA" denotes that specific language support was unavailable for that model. Here, OCR output text in red color shows that the word is incorrectly recognized. \textbf{(Best viewed in color)}. }
  \label{fig:challenges_BSTD}
\end{figure*}

\begin{table*}[!t]
\centering
\caption[Comparative evaluation of OCR systems including Google OCR, GPT-4, and our proposed IndicPhotoOCR baseline]{\textbf{Comparative evaluation for end-to-end scene text recognition on BSTD using WRR and CRR.} We present baseline results under two scenarios: (i) when text detection is provided as ground truth (denoted as oracle TD) and (ii) when both text detection and script identification are given as ground truth (denoted as TD+SI). “ \faLock/\faUnlock ” indicates
whether the tool is open-source ( \faUnlock) or closed-source (\faLock). \textbf{Bold} and \underline{underlined} number shows best performing results in non-oracle setting for closed and open-source models respectively. The asterisks (\textbf{*}) on Bengali and Devanagari indicate that Assamese and Bengali languages were merged into Bengali, and Hindi and Marathi languages were merged into Devanagari script for evaluation, owing to their shared scripts.}
\label{tab:end2end}
\renewcommand{\arraystretch}{1.3}
\resizebox{\textwidth}{!}{
\setlength{\tabcolsep}{4pt} 
\begin{tabular}{l c c c c c c c c c c c c c c}
\hline
 & \multicolumn{11}{c}{WRR on BSTD-Test} \\
 \cmidrule{2-12}
\textbf{Method} & \faLock/\faUnlock  & Bengali\textbf{*} & English & Gujarati & Devanagari\textbf{*} & Kannada & Malayalam & Odia & Punjabi & Tamil & Telugu & Average \\ \hline \hline

ChatGPT~\cite{gpt_4v}  & \faLock  &0.08	&0.36	&0.09	&0.2	&0.03	&0.05	&0.04	&0.12	&0.21	&0.1	&0.13  \\

Google OCR~\cite{googlecloud_visionocr}  & \faLock  &\textbf{0.43}	&\textbf{0.37}	&\textbf{0.32}	&\textbf{0.55}	&\textbf{0.24}	&\textbf{0.37}	&\textbf{0.43}	&\textbf{0.46}	&\textbf{0.52}	&\textbf{0.42}	&\textbf{0.41} \\

Tesseract~\cite{tesseract}   & \faUnlock  &0.04	&0.02	&0.02	&0.03	&0	&0	&0.02	&0.02	&0	&0.02	&0.02  \\

SuryaOCR~\cite{paruchuri2025surya}   & \faUnlock  &0.18	&0.15	&0.09	&0.19	&0.07	&0.12	&0.15	&0.21	&0.24	&0.05	&0.14   \\

\raisebox{-0.19\height}{\includegraphics[height=1.5em]{images/IndicPhotoOCR_LOGO.png}}  & \faUnlock &\underline{0.45}	&\underline{0.37}	&\underline{0.3}	&\underline{0.39}	&\underline{0.29}	&\underline{0.31}	&\underline{0.38}	&\underline{0.41}	&\underline{0.42}	&\underline{0.31}	&\underline{0.36} \\

\midrule

\raisebox{-0.19\height}{\includegraphics[height=1.5em]{images/IndicPhotoOCR_LOGO.png}} (+oracle TD) & \faUnlock &0.75	&0.74	&0.54	&0.63	&0.49	&0.54	&0.68	&0.7	&0.69	&0.49 &0.62\\

\raisebox{-0.19\height}{\includegraphics[height=1.5em]{images/IndicPhotoOCR_LOGO.png}} (+oracle TD+SI)   & \faUnlock &0.8	&0.92	&0.61	&0.74	&0.6	&0.58	&0.72	&0.75	&0.8	&0.56	&0.71 \\

\hline
\midrule
 & \multicolumn{11}{c}{CRR on BSTD-Test} \\
 \cmidrule{2-12}
\textbf{Method} & \faLock/\faUnlock  & Bengali\textbf{*} & English & Gujarati & Devanagari\textbf{*} & Kannada & Malayalam & Odia & Punjabi & Tamil & Telugu & Average \\ \hline \hline

ChatGPT~\cite{gpt_4v}  & \faLock &0.13	&\textbf{0.46}	&0.15	&0.28	&0.1	&0.14	&0.11	&0.2	&0.31	&0.21	&0.21 \\

Google OCR~\cite{googlecloud_visionocr}  & \faLock  &\textbf{0.57}	&0.42	&\textbf{0.36}	&\textbf{0.66}	&\textbf{0.45}	&\textbf{0.55}	&\textbf{0.57}	&\textbf{0.59}	&\textbf{0.69}	&\textbf{0.6}	&\textbf{0.55} \\

Tesseract~\cite{tesseract}   & \faUnlock  &0.12	&0.07	&0.06	&0.09	&0.05	&0.09	&0.09	&0.09	&0.04	&0.06	&0.08 \\

SuryaOCR~\cite{paruchuri2025surya}   & \faUnlock &0.29	&0.24	&0.15	&0.33	&0.19	&0.26	&0.29	&0.31	&0.45	&0.26	&0.28  \\

\raisebox{-0.19\height}{\includegraphics[height=1.5em]{images/IndicPhotoOCR_LOGO.png}}  & \faUnlock  &\underline{0.59}	&\underline{0.47}	&\underline{0.43}	&\underline{0.54}	&\underline{0.48}	&\underline{0.55}	&\underline{0.59}	&\underline{0.53}	&\underline{0.62}	&\underline{0.57}	&\underline{0.54} \\

\midrule

\raisebox{-0.19\height}{\includegraphics[height=1.5em]{images/IndicPhotoOCR_LOGO.png}} (+oracle TD) & \faUnlock  &0.87	&0.78	&0.64	&0.74	&0.68	&0.8	&0.84	&0.84	&0.78	&0.73	&0.77 \\

\raisebox{-0.19\height}{\includegraphics[height=1.5em]{images/IndicPhotoOCR_LOGO.png}} (+oracle TD+SI)   & \faUnlock &0.93	&0.97	&0.75	&0.89	&0.86	&0.86	&0.89	&0.9	&0.94	&0.83	&0.88 \\

\hline
\end{tabular}}
\end{table*}

\subsection{Results on Cropped Word Recognition}
For the cropped-word recognition baselines, there are no models specifically designed for Indian languages. Therefore, we could adopt English scene-text recognition approaches such as~\cite{clip4str, clipter, parseq, CRNN2016}. However, adopting all these methods still demands substantial computational resources, especially when running recognition models across 12 languages. Therefore, we only choose PARSeq~\cite{parseq}, one of the recent and effective recognition models, as our baseline. In addition, we also use available, CRNN~\cite{CRNN2016} by AI4Bharat~\cite{gokulkarthik_indianstr_2023} and off-the-shelf open-source models, namely Tesseract~\cite{tesseract},PaddleOCR~\cite{cui2025paddleocr30technicalreport}, and EasyOCR~\cite{jaidedai_easyocr_2020}.

PARSeq is trained on a combination of synthetic data and the BSTD dataset. Our baseline for cropped word recognition is based on the existing text recognition framework PARSeq~\cite{parseq}. We report results under two settings: (i) training PARSeq from scratch using synthetic data, and (ii) fine-tuning it on the BSTD training set. The results are presented in Table~\ref{tab:recog_results}. Fine-tuning on real data leads to substantial improvements in recognition accuracy. This highlights the critical role of real-world data in scene text recognition. Also, as can be seen, Word Recognition Rates for cropped English scene text are relatively high (around 92\%)\footnote{Unless otherwise specified, the English recognition model is trained on both uppercase and lowercase characters; however, the decoder normalizes predictions to lowercase, and evaluation is performed in a case-insensitive manner. We additionally conducted an experiment without this relaxation, i.e., using case-sensitive decoding and evaluation. Under this stricter setting, the model achieves a WRR of 84\%, compared to 92\% under case-insensitive decoding.}. However, performance drops noticeably for other languages: Marathi (86\%), Bengali (82\%), Tamil (80\%), Aasamease (79\%) Punjabi (75\%), Odia (72\%), and Hindi (71\%) form the next tier of top-performing languages. The gap widens further for low-resource languages such as Telugu and Malayalam, whose recognition accuracies remain in the 50\% range. These results clearly indicate substantial room for improving recognition performance across Indian languages. 

Futhermore, we show a selection of  visual results for several challenging cropped word images from BSTD, comparing predictions from our model against the baseline in Fig.~\ref{fig:challenges_BSTD}.


\begin{table*}[!t]
\centering
\caption[PRF metric for end2end evaluation]{\textbf{Comparative evaluation for end-to-end scene text recognition on BSTD using Precision (P), Recall (R) and F-score (F).} . The proposed baseline model, \raisebox{-0.19\height}{\includegraphics[height=1.2em]{images/IndicPhotoOCR_LOGO.png}}, significantly outperforms available open-source models and achieves competitive performance compared to commercial models. The inclusion of oracle Text Detection (TD) and Script Identification (SI) shows the potential for further improvement. The best scores for closed-source and open-source are shown in \textbf{bold} and \underline{underlined}, respectively. The asterisks (\textbf{*}) on Bengali and Devanagari indicate that Assamese and Bengali languages were merged into Bengali, and Hindi and Marathi languages were merged into Devanagari script for evaluation, owing to their shared scripts.}
\label{tab:end2end_prf}
\resizebox{\textwidth}{!}{
\begin{tabular}{lc|ccccccccccccc} 
\toprule
\multirow{2}{*}{\textbf{Method}} &
\multirow{2}{*}{\textbf{\faLock/\faUnlock}} &
\multirow{2}{*}{\textbf{Metric}} &
\multicolumn{10}{c}{\textbf{Languages}} &
\multirow{2}{*}{\textbf{Avg.}} \\
\cmidrule(lr){4-13} 
 & & & Bengali\textbf{*} & English & Gujarati & Devanagari\textbf{*} & Kannada & Malayalam & Odia & Punjabi & Tamil & Telugu & \\
\cmidrule(lr){1-14} 
\multirow{3}{*}{ChatGPT~\cite{gpt_4v}}
&\multirow{3}{*}{\faLock} 
 & P &0.11	&0.51	&0.15	&0.28	&0.05	&0.07	&0.06	&0.17	&0.23	&0.11	&0.17 \\
 & & R &0.09	&0.42	&0.12	&0.23	&0.05	&0.06	&0.06	&0.15	&0.22	&0.1	&0.15 \\
 & & F &0.09	&0.44	&0.13	&0.24	&0.05	&0.06	&0.05	&0.15	&0.22	&0.1	&0.15 \\
 
\cmidrule(lr){2-14}
\multirow{3}{*}{Google OCR~\cite{googlecloud_visionocr}}
&\multirow{3}{*}{\faLock}
 & P & \textbf{0.52}	 &\textbf{0.54}	&\textbf{0.42}	&\textbf{0.64}	&\textbf{0.34}	&\textbf{0.45}	&\textbf{0.54}	&\textbf{0.59}	&\textbf{0.58}	&\textbf{0.49}	&\textbf{0.51} \\
 & & R &\textbf{0.55}	&\textbf{0.71}	&\textbf{0.44}	&\textbf{0.65}	&\textbf{0.3}	&\textbf{0.42}	&\textbf{0.61}	&\textbf{0.6}	&\textbf{0.64}	&\textbf{0.53}	&\textbf{0.54} \\
 & & F &\textbf{0.52}	&\textbf{0.59}	&\textbf{0.42}	&\textbf{0.63}	&\textbf{0.31}	&\textbf{0.43}	&\textbf{0.56}	&\textbf{0.58}	&\textbf{0.6}	&\textbf{0.5}	&\textbf{0.51} \\
 
\cmidrule(lr){2-14}
\multirow{3}{*}{Tesseract~\cite{tesseract}}
&\multirow{3}{*}{\faUnlock}
 & P &0.08	&0.05	&0.05	&0.06	&0.01	&0	&0.04	&0.08	&0	&0.03	&0.04 \\
 & & R &0.08	&0.11	&0.03	&0.05	&0.02	&0.01	&0.04	&0.06	&0.01	&0.02	&0.04 \\
 & & F &0.07	&0.06	&0.03	&0.05	&0.01	&0.01	&0.03	&0.05	&0	&0.02	&0.03 \\
 
\cmidrule(lr){2-14}
\multirow{3}{*}{SuryaOCR~\cite{paruchuri2025surya}}
&\multirow{3}{*}{\faUnlock}
 & P &0.27	&0.3	&0.19	&0.29	&0.11	&0.17	&0.23	&0.38	&0.34	&0.1	&0.24 \\
 & & R &0.24	&0.36	&0.1	&0.23	&0.09	&0.15	&0.18	&0.24	&0.29	&0.13	&0.20 \\
 & & F &0.23	&0.29	&0.12	&0.25	&0.09	&0.15	&0.19	&0.28	&0.29	&0.11	&0.2 \\
 
\cmidrule(lr){2-14}
\multirow{3}{*}{\raisebox{-0.19\height}{\includegraphics[height=1.5em]{images/IndicPhotoOCR_LOGO.png}}}
&\multirow{3}{*}[-3pt]{\faUnlock} 
 & P &\underline{0.59}	&\underline{0.55}	&\underline{0.43}	&\underline{0.53}	&\underline{0.41}	&\underline{0.39}	&\underline{0.47}	&\underline{0.56}	&\underline{0.55}	&\underline{0.38}	&\underline{0.49} \\
 & & R &\underline{0.53}	&\underline{0.46}	&\underline{0.4}	&\underline{0.43}	&\underline{0.33}	&\underline{0.38}	&\underline{0.47}	&\underline{0.49}	&\underline{0.52}	&\underline{0.37}	&\underline{0.44} \\
 & & F &\underline{0.55}	&\underline{0.48}	&\underline{0.4}	&\underline{0.46}	&\underline{0.36}	&\underline{0.38}	&\underline{0.47}	&\underline{0.51}	&\underline{0.52}	&\underline{0.37}	&\underline{0.45} \\
 
\midrule
\multirow{3}{*}{\raisebox{-0.19\height}{\includegraphics[height=1.5em]{images/IndicPhotoOCR_LOGO.png}}(+oracle TD)}
&\multirow{3}{*}[-3pt]{\faUnlock} 
 & P &0.72	&0.83	&0.65	&0.66	&0.56	&0.57	&0.65	&0.69	&0.78	&0.54	&0.66 \\
 & & R &0.69	&0.75	&0.62	&0.61	&0.5	&0.57	&0.63	&0.68	&0.73	&0.53	&0.63 \\
 & & F &0.7	&0.78	&0.62	&0.62	&0.52	&0.57	&0.63	&0.68	&0.75	&0.53	&0.64 \\
\cmidrule(lr){2-14}
\multirow{3}{*}{\raisebox{-0.19\height}{\includegraphics[height=1.5em]{images/IndicPhotoOCR_LOGO.png}}(+oracle TD+SI)}
&\multirow{3}{*}[-3pt]{\faUnlock}
 & P &0.75	&0.89	&0.71	&0.71	&0.57	&0.6	&0.66	&0.72	&0.8	&0.55	&0.70 \\
 & & R &0.75	&0.9	&0.72	&0.71	&0.56	&0.59	&0.66	&0.73	&0.81	&0.56	&0.70 \\
 & & F &0.75	&0.89	&0.71	&0.71	&0.56	&0.6	&0.66	&0.72	&0.8	&0.56	&0.70 \\
\bottomrule
\end{tabular}}
\end{table*}

\subsection{End-to-End Evaluation}
\label{sec:end-2-end}
We now present end-to-end evaluation results on the BSTD benchmark. For this, we compare our baseline with the following methods:

\noindent\textbf{1. Closed-Source Commercial Baselines:}
We evaluated two commercial systems, namely GPT-4 with vision~\cite{gpt_4v} and Google OCR~\cite{googlecloud_visionocr} on the BSTD benchmark\footnote{We obtained their results using their best available API on October 31, 2025.}. Although these comparisons are not entirely fair due to the opaque nature of their training data and model architectures, they offer a useful reference point to gage the current state of the art in Indian language scene text recognition.

We leveraged OpenAI’s GPT-4 with Vision (accessed via the \texttt{gpt-4o} model) to perform zero-shot scene text recognition directly. We use the following structured prompt: \textit{``Extract all text from this image. Return all words in a list format like this: [`word1', `word2']"}, with a system message defining the model's role as: \textit{``You are an OCR system that extracts text from images."} 
Although with more detailed prompt, such as asking to locate text or perform chain-of-thoughts reasoning, this baseline may be improved. However, such prompt engineering is beyond the scope of this paper. 

Further, we also utilized the Google Cloud Vision API to perform OCR on BSTD (Test Set).

\begin{figure*}[!t]
    \centering
    \includegraphics[width=16cm]{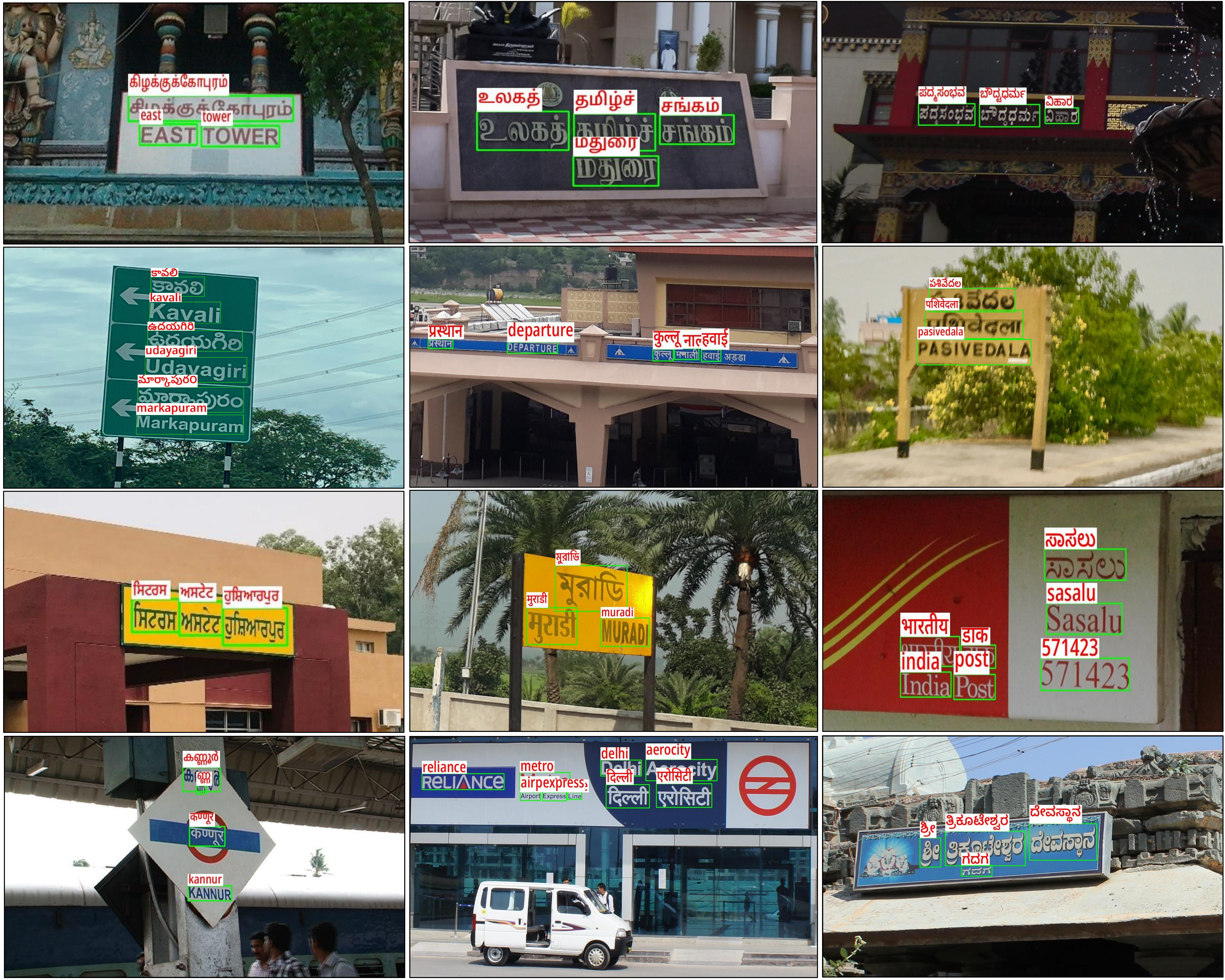}
    \caption[End-2-End results from IndicPhotoOCR]{A selection of results using \raisebox{-0.19\height}{\includegraphics[height=1.5em]{images/IndicPhotoOCR_LOGO.png}}. Detected bounding boxes are shown in green, with the recognized text displayed in red on a white background adjacent to each box. \textbf{(Best viewed in color)}.}
    \label{fig:end2end_image_results}
\end{figure*}
\begin{figure*}[!t]\centering
    \includegraphics[width=14cm]{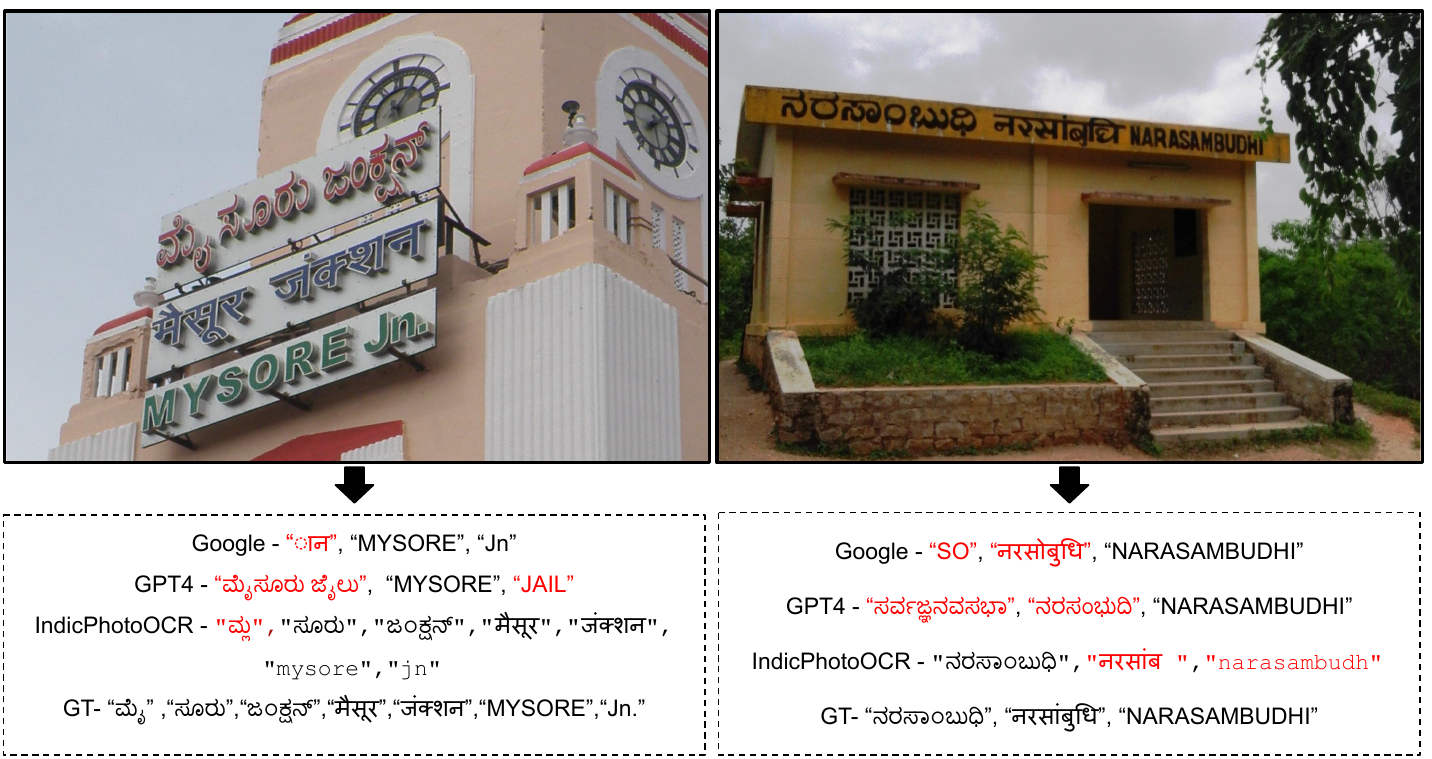}
    \caption[Visual Error Analysis of Google, ChatGPT and IndicPhotoOCR]{
    A couple of challenging examples with output text by closed source models and ours. Text in red color shows incorrect recognition. \textbf{(Best viewed in color)}.}
    
    \label{fig:gpt_google_fails}
\end{figure*}

\noindent\textbf{2. Open-Source Baselines:}
Currently, apart from our own, no open-source system is explicitly designed for Indian language scene text recognition. However, we include comparisons with general-purpose OCR tools such as Tesseract~\cite{tesseract} and SuryaOCR~\cite{paruchuri2025surya}, which are widely used in practical settings. 

To further analyze the performance of \raisebox{-0.19\height}{\includegraphics[height=1.2em]{images/IndicPhotoOCR_LOGO.png}}, we also introduce the following oracle variants: (i) \textit{IndicPhotoOCR (+Oracle TD)}: Uses ground-truth text detection. (ii) \textit{IndicPhotoOCR (+Oracle TD + Oracle SI)}: Uses ground-truth text detection and script identification. These variants help estimate the upper bound on performance in scenarios where text detection or both text detection and script identification are solved reliably.

Comparison with the aforementioned competitive baselines is presented in Table~\ref{tab:end2end} and Table~\ref{tab:end2end_prf} using WRR,CRR and Precision/Recall and F1 score. We observe that \raisebox{-0.19\height}{\includegraphics[height=1.2em]{images/IndicPhotoOCR_LOGO.png}} consistently outperforms all existing open-source models across all the evaluated languages. Remarkably, it also surpasses the popular commercial vision-language model GPT-4, which may not be optimized for Indian scripts. However, our approach still lags behind Google OCR in some cases, highlighting a performance gap that we aim to address in future work. Encouragingly, the variant IndicPhotoOCR (+oracle TD + SI) achieves results that are competitive, infact 30\% (WRR) superior to Google OCR on an absolute scale on average, indicating the potential of our pipeline with improved text detection and script identification.  In Table~\ref{tab:end2end_prf}, on P/R/F metric, \raisebox{-0.19\height}{\includegraphics[height=1.2em]{images/IndicPhotoOCR_LOGO.png}} significantly surpasses other open-source baselines. The proposed baseline also not significantly behind the commercial state-of-the-art Google OCR. In fact, when combined with oracle text detection and oracle script identification, our baseline can significantly outperform Google OCR, highlighting its potential for further improvement.

A somewhat surprising observation is the noticeable performance drop for English in the end-to-end setting. Although the proposed baseline achieves a cropped word recognition accuracy of 92\%, its end-to-end performance declines sharply. This drop is primarily attributable to script identification and text detection errors introduced in the full pipeline. Notably, this pattern is also reflected in commercial OCR systems such as Google’s OCR.

\begin{figure*}[!t]
    \centering
    \includegraphics[width=16cm]{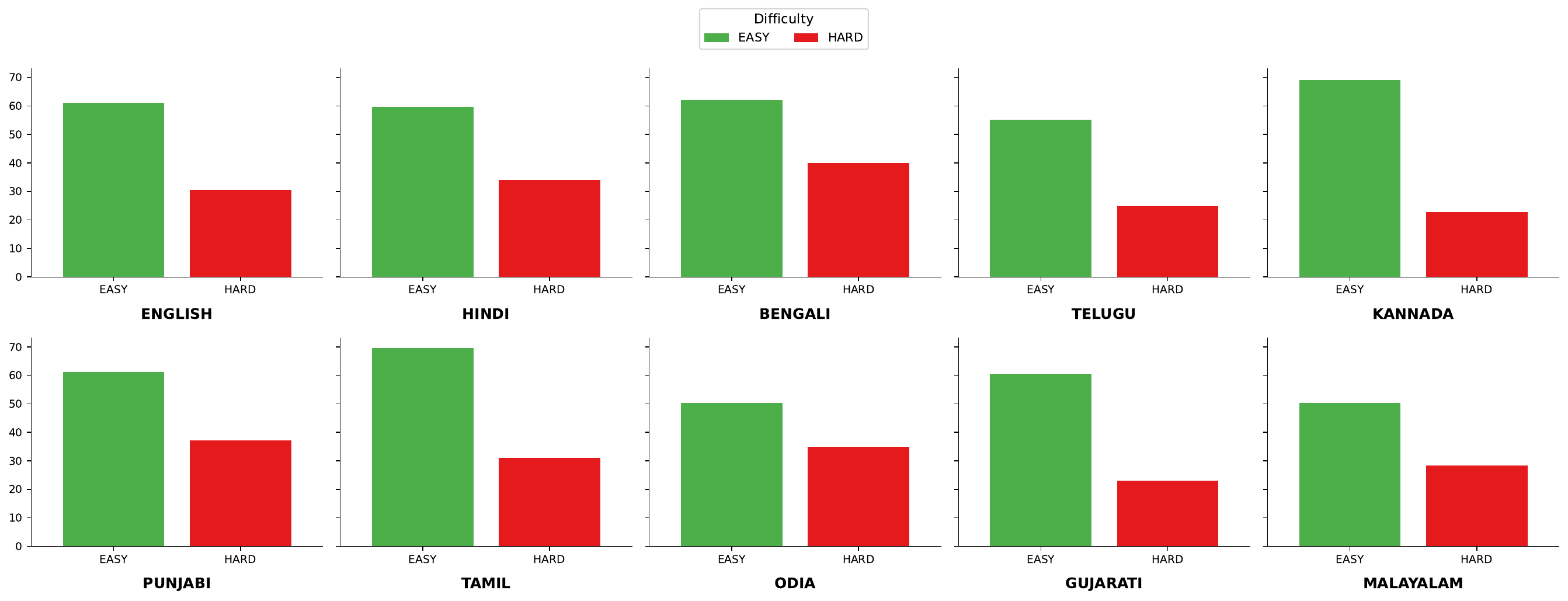}
    \caption[IndicPhotoOCR WRR by Language and Difficulty]{Average Word Recognition Rate (WRR) based on difficulty level. \textbf{(Best viewed
in color)}.}
    \label{fig:wrr_languagewise_difficulty}
\end{figure*}

We also performed a comprehensive qualitative analysis of the proposed baseline. A selection of results is shown in Fig.~\ref{fig:end2end_image_results}. We observe that \raisebox{-0.19\height}{\includegraphics[height=1.2em]{images/IndicPhotoOCR_LOGO.png}} demonstrates strong multilingual scene text recognition across diverse real-world settings - from street signs and transportation hubs to institutional boards and cultural heritage sites. The model performs well even on slanted or perspective-distorted text, e.g., directional signs, railway nameplates, as well as on low-resolution or distant text, indicating robustness in challenging visual conditions. Further, we show a couple of challenging examples in Fig.~\ref{fig:gpt_google_fails} where even closed-source models, including ours, fail to recognize the text correctly. 

\begin{figure*}[!t]
    \centering
    \includegraphics[width=16cm]{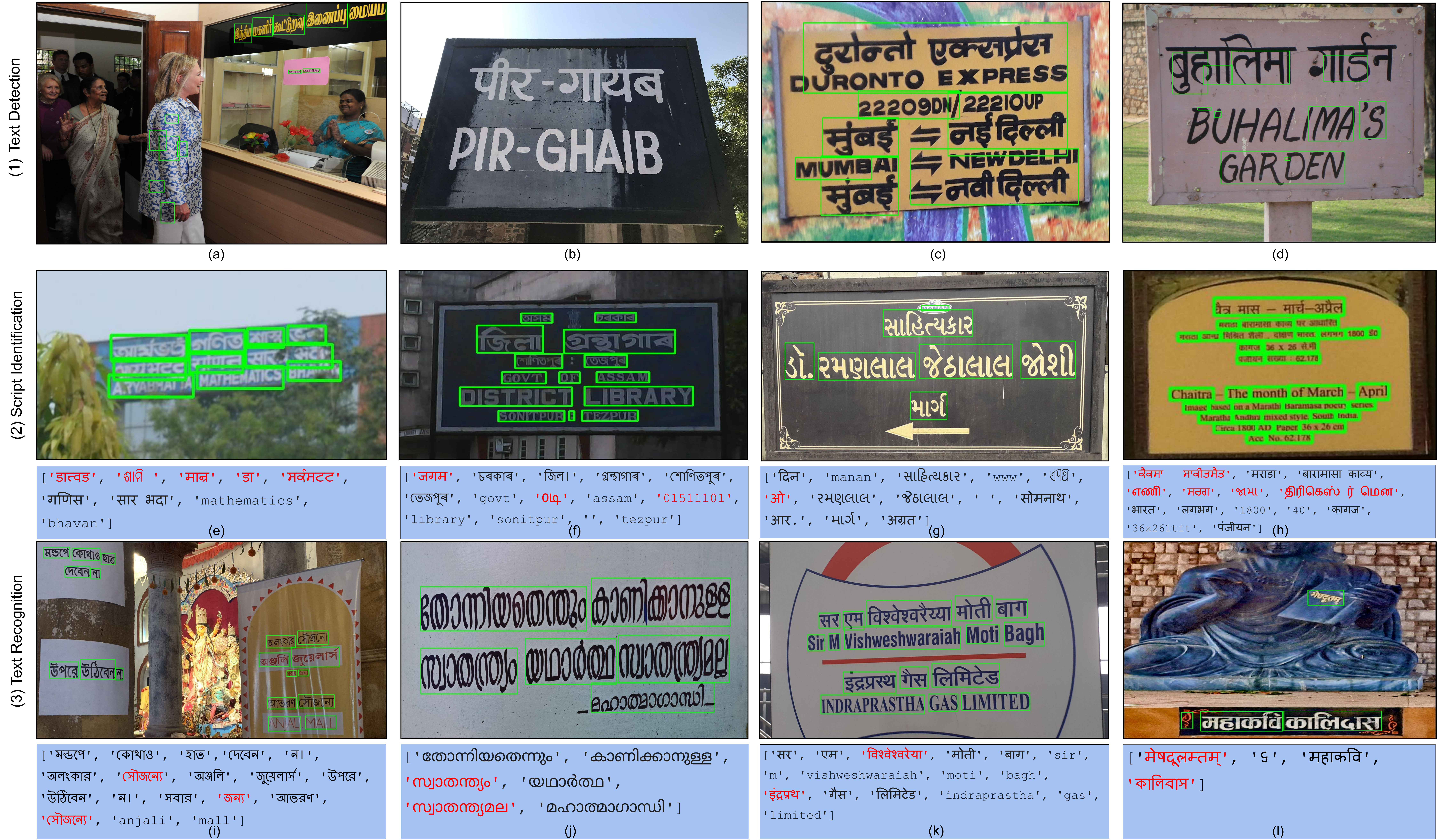}
    \caption{Failure cases in end-to-end scene text processing: The images showcase representative errors observed across the pipeline. (1) Text detection failures from TextBPN++, including false positives, missed detections, merged text, and large text misdetections. (2) Script identification errors from the ViT-based model, highlighting misclassified scripts, multi-script confusion, and challenges with degraded text. (3) Recognition failures from PARSeq, demonstrating incorrect predictions, missing or hallucinated characters, and cross-script misinterpretations. \textbf{(Best viewed
in color)}.}
    \label{fig:end-2-end-failures}
\end{figure*}

On the BSTD test set, we additionally asked annotators to label each image as easy or hard based on the readability of the scene text. In the easy subset, the proposed \raisebox{-0.19\height}{\includegraphics[height=1.5em]{images/IndicPhotoOCR_LOGO.png}} achieves 60\% WRR, whereas in the hard subset the performance drops to 31\% across all 10 scripts. Script-wise word recognition rates are presented in Fig.~\ref{fig:wrr_languagewise_difficulty}. This clear separation between easy and hard cases highlights the sensitivity of current methods to challenging visual conditions and calls for more robust recognition models.

\subsection{Error Analysis}
The proposed baseline is not flawless and leaves several avenues for improvement. We performed an extensive error analysis of each module of our pipeline (Section~\ref{sec:end-2-end}), summarized in Fig.~\ref{fig:end-2-end-failures}, and highlight the main failure modes below.

\noindent\textbf{Text Detection}: Common issues in text detection include (i) \textit{false positives} where background edges resemble text, (ii) \textit{missed detections} in low-contrast regions, (iii) \textit{merged detections} when nearby words are grouped together, and (iv) difficulty with \textit{extremely large text}, suggesting the model is tuned more for small or medium size words.

\noindent\textbf{Script Identification}: (i) \textit{Misclassifications} occur among visually similar scripts, e.g., Gujarati-Devanagari, Bengali-Assamese. (ii) \textit{Multilingual overlap} leads to errors when multiple scripts co-occur, as the model assumes a single script per region. (iii) \textit{Degraded or stylized text} often confuses the classifier due to missing or distorted characters.

\noindent\textbf{Text Recognition}: Errors include (i) \textit{incorrect predictions} in complex scripts like Telugu or Tamil, (ii) \textit{hallucinated or missing characters}, and (iii) failures on \textit{eroded/unclear text}.

These findings emphasize the need for cross-module optimization and feedback-driven refinement between detection, script identification, and recognition. In addition, these qualitative as well as the quantitative analysis in the earlier section, suggest that improving each module in the pipeline can significantly improve overall performance.

\section{Conclusion and Future Work}
In this work, we introduced the Bharat Scene Text Dataset (BSTD), a large-scale benchmark specifically designed to address the gap in Indian language scene text recognition. BSTD comprises 6,582 scene images containing 1,26,292 words across 11 Indian languages and English, making it the most comprehensive dataset of its kind. Through rigorous manual annotation, we ensured high-quality polygon-level bounding boxes and text transcriptions. We also provided extensive benchmarks using state-of-the-art text detection, script identification, cropped word recognition, and end-to-end scene text recognition methods, offering a strong foundation for future research.

Our evaluations highlight the challenges posed by Indian scripts, including complex character structures, diverse vocabulary, and script variations. The results also demonstrate that existing models, primarily designed for English and Latin-based scripts, struggle with Indian languages, emphasizing the need for improved recognition techniques. Despite our extensive efforts, several challenges remain open for exploration in the future. We list them here: 

\noindent\textbf{Dataset Expansion:} While BSTD provides a strong foundation, future extensions could include more languages (especially Urdu and Meiti) and larger-scale data to improve generalization.

\noindent\textbf{Improved Recognition Models:} Developing novel architectures that better capture the complexities of Indian scripts, including joint modeling of script identification and recognition, remains an open research avenue.

\noindent\textbf{Benchmarking and Community Involvement:} We encourage the research community to contribute to BSTD by proposing new models, sharing additional annotations, and utilizing the dataset for novel applications such as scene text translation, Indian language scene text aware captioning and question answering, and cross-lingual scene text-aware image retrieval.

We firmly believe that the BSTD and open-source toolkit \raisebox{-0.19\height}{\includegraphics[height=1.3em]{images/IndicPhotoOCR_LOGO.png}} will serve as catalyst for future advancements in the Indian language scene-text recognition and associated problems, inspiring researchers to build robust and scalable solutions for multilingual photoOCR.

\section*{Acknowledgments}
This work is supported by the Ministry of Electronics and Information Technology, Govt. of India, under the NLTM-Bhashini Project. 

\backmatter

\small
\bibliography{sn-bibliography}

\newpage
\begin{appendices}
\clearpage
\section*{Appendix A \vspace{1em} \\ \raisebox{-0.19\height}{\includegraphics[height=1.5em]{images/IndicPhotoOCR_LOGO.png}}: Installation and Usage}
\addcontentsline{toc}{section}{Appendix A: IndicPhotoOCR}
\label{sec:indicphotoocr_appendix}
We open-sourced \raisebox{-0.19\height}{\includegraphics[height=1.4em]{images/IndicPhotoOCR_LOGO.png}}, a toolkit for recognizing text in English and 11 Indian languages: Assamese, Bengali, Gujarati, Hindi, Kannada, Malayalam, Marathi, Odia, Punjabi, Tamil, and Telugu.\footnote{Future versions will include Urdu and Meitei.} This toolkit is designed to handle the unique scripts and complex layouts of Indian languages. By releasing this toolkit publicly, we enable reproducible research, provide a common benchmark for multilingual scene text, and support the development of real-world applications across diverse Indian scripts.

We provide a user-friendly installation and usage platform, allowing our proposed baseline to be easily replicated. We firmly believe that this open-source effort will accelerate future research in Indian language scene text recognition. The installation and usage instructions are provided below.

\subsection*{Installation}
\begin{bashbox}{Terminal - Setup}
$ git clone https://github.com/Bhashini-IITJ/IndicPhotoOCR.git
$ cd IndicPhotoOCR
$ chmod +x setup.sh
$ ./setup.sh
\end{bashbox}

\subsection*{Usage}
\begin{pythonbox}{Text Detection}
from IndicPhotoOCR.ocr import OCR

# Initialize the OCR system
ocr_system = OCR(verbose=True)

# Define input image path
image_path = "path/to/image.jpg"

# Perform text detection
detections = ocr_system.detect(image_path)

# Save and visualize the detection results
ocr_system.visualize_detection(image_path, detections)
# Output image saved
\end{pythonbox}

\begin{pythonbox}{Script Identification}
from IndicPhotoOCR.ocr import OCR

# Initialize the OCR system with automatic language identification
ocr_system = OCR(identifier_lang="auto")
image_path = "path/to/cropped_image.jpg"

# Perform script identification
recognized_lang = ocr_system.identify(image_path)

# Output recognized words
print(recognized_lang)
\end{pythonbox}

\begin{pythonbox}{Cropped Word Recognition}
from IndicPhotoOCR.ocr import OCR

# Initialize the OCR system
ocr_system = OCR(verbose=True)
image_path = "path/to/cropped_image.jpg"
language = "hindi"

# Perform text recognition
recognized_text = ocr_system.recognise(image_path, language)

# Output recognized text
print(recognized_text)
\end{pythonbox}

\begin{pythonbox}{End-to-End Scene Text Recognition}
from IndicPhotoOCR.ocr import OCR

# Initialize the OCR system with automatic language identification
ocr_system = OCR(identifier_lang="auto")

# Define input image path
image_path = "path/to/image.jpg"

# Perform end-to-end OCR
recognized_words = ocr_system.ocr(image_path)

# Output recognized words
print(recognized_words)
\end{pythonbox}

The tool is reasonably lightweight; the detection module consists of 30M parameters, the script identification module contains 80M parameters, and the 12 recognition modules (23M each) aggregate to 386M parameters in total. This architecture offers a viable alternative to large vision-language-based OCR systems, particularly for deployment in resource-constrained and on-device computing environments.
Detailed documentation for usage of our toolkit is available here\footnote{\url{https://bhashini-iitj.github.io/IndicPhotoOCR/}}.
\end{appendices}
\end{document}